\documentclass[preprint,12pt]{elsarticle}

\usepackage{graphicx}
\usepackage{amssymb}

\usepackage{lineno}

\usepackage[utf8]{inputenc}
\usepackage[T1]{fontenc}
\usepackage{algorithm}
\usepackage{multirow}
\usepackage[noend]{algpseudocode}
\usepackage{enumitem,linegoal}
\newtheorem{HR1}{Heuristic}
\usepackage{tikz}
\usetikzlibrary{shadows.blur,positioning,calc,arrows.meta}
\tikzstyle{pinkbox}=[draw,rounded corners,shade,top color=white,bottom color=pink,
minimum width=3.5cm,minimum height=1cm,align=center,node distance=1.5cm]
\tikzstyle{inter arrow}=[->,ultra thick,-{Triangle[angle=45:8pt]}]

\journal{Information Sciences}

\begin{document}

\begin{frontmatter}


\title{Automatic topography of high-dimensional data sets by non-parametric Density Peak clustering}


\author[1,2,3]{Maria d'Errico}%
\author[1]{Elena Facco}%
\author[1,4]{Alessandro Laio}
\ead{laio@sissa.it}
\author[1,4]{Alex Rodriguez}
\ead{arodrigu@ictp.it}
\address[1]{ SISSA, Scuola Internazionale Superiore Studi Avanzati,  via Bonomea 265, I-34136 Trieste, Italy}%
\address[2]{ Functional Genomics Center Z{u}rich, UZH/ETHZ,  8057 Z{u}rich, Switzerland}%
\address[3]{ Swiss Institute of Bioinformatics, Quartier Sorge - Batiment Amphipole  1015 Lausanne, Switzerland}%
\address[4]{ICTP, International Centre for Theoretical Physics,  Strada Costiera 11,I-34100, Trieste, Italy}


\begin{abstract}
Data analysis in high-dimensional spaces aims at obtaining a synthetic description of a data set, revealing its main structure and its salient features.
We here introduce an approach providing  this description in the form of a \emph{topography} of the data, namely a human-readable chart of the probability density from which the data are harvested.
The approach is based on an unsupervised extension of Density Peak clustering and a non-parametric density estimator that measures the probability density in the manifold containing the data. This allows finding automatically the number and the height of the  peaks of the probability density, and the depth of the ``valleys'' separating them.  Importantly, the density estimator provides a measure of the error, which allows distinguishing genuine density peaks from density fluctuations due to finite sampling.
The approach thus provides robust and visual information  about the density peaks height, their statistical reliability and their hierarchical organization, offering a conceptually powerful extension of the standard clustering partitions. We show that this framework is  particularly useful in the analysis of complex data sets.

\end{abstract}

\begin{keyword}
Clustering-algorithm\sep High-dimensional-data\sep Hierarchy-visualization\sep density-peak-clustering\sep Non-parametric-density-estimation


\end{keyword}

\end{frontmatter}


\section{Introduction}
\label{S:1}
The rapidly increasing capability to generate data calls for 
approaches able to provide a compact representation of their underlying structure.
The challenge is to extract from data sets with, say, 1000 dimensions an information content at the same time human readable and useful. 

A possible route to achieve this goal is attempting to map the data
on a two or three dimensional surface that can then be directly visualized.
{This low-dimensional representations of high-dimensional data can be derived, for example, using}  Principal Component Analysis~\cite{PCA} and,
within a framework which allows taking non-linearities into account, in multidimensional scaling~\cite{torgerson1952multidimensional}, ISOMAP~\cite{tenenbaum2000global}, Diffusion Maps~\cite{DM}, Locally Linear Embedding~\cite{LLE}, Tree Preserving Embedding~\cite{TPE}, t-Distributed Stochastic Neighbor Embedding~\cite{tsne} and Sketch-Map~\cite{SKETCHMAP}.
However, the intrinsic dimensionality (ID) of realistic data sets is often larger than three.
This has become more and more evident in recent years, thanks to the development of powerful
and accurate approaches capable of estimating the ID~\cite{CAMASTRA201626,granata2016accurate,faccoID}.
If the ID of a data set is, say, 10 any attempt to describe it with only two or three
coordinates unavoidably leads to an information loss.
{This can lead to several effects like altering the local neighborhood structures of the data sets, or the crowding of the data due to the reduction of space when passing from high dimension to low dimension~\cite{gisbrecht2015data}. Thus, most of the above-mentioned methods try to quantify the amount of information preserved in the projection by a suitable objective function. For instance, in Multidimensional Scaling the objective function measures the preservation of the distances, while in t-SNE the information preserved is the neighborhood structure~\cite{bunte2012general}.
However, this information loss can lead, depending on the data set, to the presence of artifacts that make those methods not applicable for a quantitative analysis in case of large IDs, although they may be still useful as preprocessing, or for visualization purposes. One example (among many) of these problems can be found in ref.~\cite{sormani2019explicit}, where ISOMAP was employed for projecting a folding trajectory of Villin protein described by the 32 features (with an ID of 12). In this particular case the projection was not able of distinguish folded or unfolded states nor provide a useful visualization of the data set.}

A different strategy for summarizing the information content of a data set is considering the
data as an ensemble of realizations drawn from a probability distribution
{where the regions of space with higher density of data points, generally defined as clusters, correspond to probability peaks.} 
{Density-based clustering~\cite{corrected_DBSCAN,MeanShift2004,AlexAle} allows finding those peaks and estimating their properties without projecting the data onto a lower-dimensional representation}.  
{This approach offers two major advantages. First, it can be followed even in the case of high intrinsic dimensionalities. Second, it can also  be exploited to formulate a hierarchical representation of the
probability distribution, by establishing a hierarchy of connected subsets of points}.
This idea was introduced  in the seminal work of Hartigan~\cite{hartigan1981consistency}, and has been exploited in many recent algorithms
like, for instance, HDBSCAN~\cite{HDBSCAN}, Robust Single Linkage~\cite{RobustSL} and
Robust Density-Based Clustering~\cite{RobustDB}.

In this work we introduce Density Peaks Advanced (DPA), a method for reconstructing what we call the \emph{topography} of a
data set, or a simplified human-readable chart of the probability distribution. The topography conveys information on
the height of all the probability peaks as well as on the organization of these peaks in larger structures. When density peaks are identified as clusters, the topography provides an immediate visual information about their relationships to one another.
If the probability distribution includes $N$ peaks (or clusters), the topography consists in a $N\times N$
symmetric matrix in which the diagonal entries are the heights of the peaks and the off-
diagonal entries are the heights of the saddle points. 
As we discuss in the following, a saddle point between two peaks is estimated by searching the point of highest density among all the points at the border between the two peaks. 
An off-diagonal entry is
set to zero if the two peaks are not in contact. This matrix can be represented in the form of
a tree diagram, like in refs.~\cite{hartigan1981consistency,HDBSCAN,RobustSL,RobustDB}, 
obtaining a chart that unveils the hierarchies by focusing on the highest saddles
between peaks. We will also show that complementary information about the topography can be visualized by applying one of
the approaches developed for representing the kinetic models derived by Markov State Model
analysis~\cite{MSM2010}.

The topography is reconstructed by using a modified version of the Density Peaks (DP) algorithm~\cite{AlexAle}.
This approach provides an empirical criterion for a quick and reliable lo\-ca\-li\-za\-tion of density peaks. The original formulation of DP is affected by two main drawbacks. First, the selection of
cluster centers is relatively subjective, since it is
based on the visual inspection of the so-called decision graph: a scatter plot of the density of a point vs its
minimum distance  from a point with higher density (see ref.~\cite{AlexAle}).
Second, like all density-based clustering approaches, it is sensitive to the parameters
involved in the density estimation~\cite{Xu2015}. These drawbacks have been addressed in many works.
For instance, ref.~\cite{LIANG201652} automatically finds the number of clusters through a recursive
inspection of the decision graph based on a Divide-and-Conquer strategy.
In the two-dimensional case, ref.~\cite{MEHMOOD} introduces instead a non-parametric technique for estimating the
densities based on the heat diffusion equation. Although the method in ref.~\cite{MEHMOOD} still requires the inspection of
the decision graph, it shows an improved performance
in the classification of artificial data sets.

In this work we demonstrate that DP clustering algorithm can be made fully unsupervised 
by combining it with  PA$k$~\cite{FEwithoutCV}, a non-parametric density  estimator
recently proposed by us. {This estimator is able to exploit a statistical approach to find
the largest region around each point in which the density is approximately
constant.} 
One of the PA$k$'s main innovations with respect to other non-parametric estimators is that it
measures the density in the manifold in which the data lay, and not in the embedding space whose dimensionality is normally overwhelmingly large.
In the following we show that the ma\-the\-ma\-ti\-cal formulation of the estimator in ref.~\cite{FEwithoutCV}
naturally induces a criterion to automatically find density peaks through the DP clustering.
Moreover, the estimator provides a measure of the uncertainty on the density. This last feature is a key ingredient, since it allows recognizing genuine
density peaks from statistical fluctuations of the estimated density due to finite sampling. In fact, our approach allows 
assessing the statistical reliability of probability peaks thus providing a new manner for performing a so-called multimodality test (see, for instance, refs.~\cite{silverman1981using,minnotte1997nonparametric}).
This, to the best of our knowledge, is an original contribution of this work.

Finally, after testing the DPA algorithm in several toy problems, we analyze two real world data sets: the MNIST database~\cite{MNIST} of handwritten digits and a sample of protein sequences extracted from the Pfam clan PUA~\cite{PFAM,PFAMlatest}, a complex superfamily of sequences organized into ten families, each containing a variety of architectures.

\section{Methods}

\subsection{\label{sec:Topography}Topography of probability density function landscapes}
\begin{figure}
        \centerline{\includegraphics[width=0.7\linewidth]{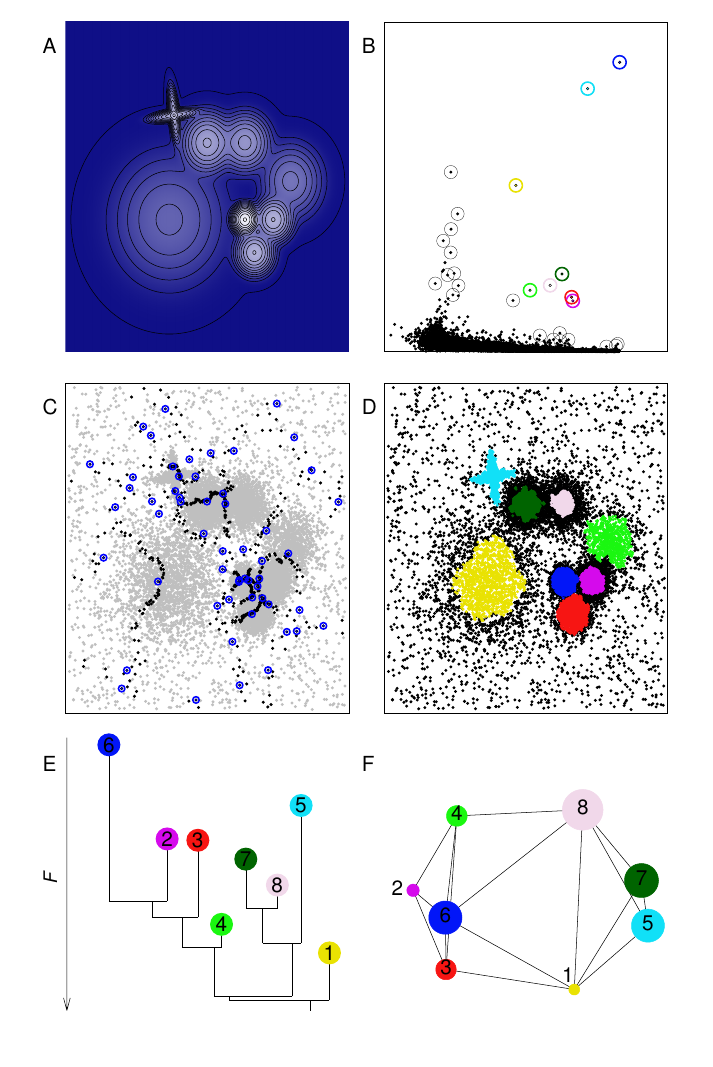}}
        \caption{CLUS8, a two dimensional toy example. (A) Probability density function from which the 20000
    points have been drawn. (B) Decision graph using $g$ as density. Circled points correspond
    to the putative centers chosen according with Heuristic~\ref{hr1}, while the colored ones are the centers of the ground truth clusters. 
     (C) Borders (black points) and saddles (blue points) between clusters. (D) Final assignation of
    the points to clusters, black points correspond to halo points. The color code is the same for
    Panels B, E and F. (E) Dendrogram representation of the data set topography. (F) Network
    representation of the data set topography. }\label{fig:Clusterexpl}
\end{figure}

Data sets can often be described as realizations of an underlying probability distribution whose density has support 
 in the space of the features (coordinates) of the data.
This density can be characterized by the presence of several maxima, at times organized hierarchically, as in the synthetic example shown in Figure~\ref{fig:Clusterexpl}A.

The scope of our approach is reconstructing {with no supervision}
the topography of {those complex} probability distributions.
A key step of this procedure consists in the identification of the density
maxima within the data set and the saddle points between them.

We will show that the density peaks and saddle points can be
automatically recognized by making use of the PA$k$ {(Point Adaptive $k$-nearest neighbor)} density estimator~\cite{FEwithoutCV}
within the framework of Density Peaks clustering~\cite{AlexAle}.

\subsubsection{\label{sec:PaK}An adaptive $k$-Nearest Neighbor density estimator}

In this section we will describe in short the PA$k$ density estimator as introduced in ref.~\cite{FEwithoutCV}.
PA$k$ aims to estimate the local density around each point $i$ in a data set.
We denote by $\{r_{i,l}\}_{l \le k}$  the sequence of the ordered distances between $i$ and its first
$k$-nearest neighbors, and by $v_{i,l}=\omega\left(r_{i,l}^{d}-r_{i,l-1}^{d}\right)$ 
the volumes of the hyperspherical shells enclosed between two successive neighbors $l-1$ and $l$,
where $d$ is the Intrinsic Dimension (ID) of the manifold in which the
data points lay. As we discussed in ref.~\cite{FEwithoutCV} the density should be estimated by measuring the volumes in the embedding manifold whose dimension is $d$,
rather than the extrinsic dimension of the space in which the data points are defined, which can be orders of magnitude larger than $d$. The value of $d$ can be estimated using one of the many approaches for computing the intrinsic dimension. We here estimate it by the TWO-NN method~\cite{faccoID}.

PA$k$ relies on the observation that, for a given point $i$, if the density is constant the volumes $v_{i,l}$ are independently drawn from an exponential distribution with rate equal to the density $\rho$, as proven in ref.~\cite{faccoID}. 
Therefore, the log-likelihood function of the parameter $\rho$ given the observation of
the $k$-nearest neighbor distances from point $i$ is
\begin{equation}
\mathcal{L}_{i,k}\left(\rho_i\right)=k\cdot\log\rho_i-\rho_i\cdot\sum_{l=1}^{k}v_{i,l}=k\cdot\log\rho_i-\rho_i  \cdot V_{i,k}. \label{eq:logML}
\end{equation}
where $V_{i,k}$ denotes the total volume occupied by the $k$ nearest neighbors of point $i$.

Indeed, the maximization of equation~\ref{eq:logML} with respect $\rho_i$ leads to the $k$-Nearest Neighbor density estimator with an associated variance that decreases with the square root of $k$,
so the estimate improves when using high values of $k$.
However, when $k$ increases, the density in the neighborhood within a distance $r_{i,k}$ 
from the data point $i$ can become non-constant, breaking the hypothesis from which equation~\ref{eq:logML} has been derived and therefore inducing systematic errors (bias).
PA$k$ finds, for each data point, the largest neighborhood $\hat{k}$ at which the density can be considered constant by performing a Likelihood Ratio
Test~\cite{LikeRat} between two models: (1) {a model} in which the densities at the point and at its $k^{th}$ nearest neighbor are considered the same and (2) a model in which the densities are assumed to be different.
This test 
involves estimating the difference $D_k$ of the log-likelihood of the two models for increasing values of $k$.
For each point $i$ the algorithm chooses the optimal value of $k$, denoted by $\hat{k}$, according to the condition
\begin{equation}
        \hat{k}_i: (D_k < D_{\textrm{thr}}\ \forall\  k \le \hat{k}_i)  \& (D_{\hat{k}_i+1} \ge D_{\textrm{thr}})
\end{equation}
where $D_{\textrm{thr}}=23.928$, corresponding to  a p-value of  10$^{-6}$. 
This implies that for point $i$ the log-likelihoods of the two models are consistent within a \mbox{p-value of 10$^{-6}$} 
for  $k \le \hat{k_i}$.

Once the optimal $k$ for point $i$ {has been found}, the algorithm corrects for the systematic errors
induced by choosing $\hat{k_i}$ according to a fixed confidence threshold by modifying the log-likelihood function in equation~(\ref{eq:logML}) to include an extra
variational parameter $a$. {This parameter describes the linear trend in the density estimation as one moves further and further
from the central point}. {At this point} the  density $\rho_i$ and the corresponding uncertainty $\epsilon_i$ are
obtained by maximizing the log-likelihood in equation~(\ref{eq:logML}) with $k$ equal to $\hat{k}_i$ and $\log{(\rho)}$
replaced by $\log{(\rho)}+a\cdot l$.
By {leveraging} Fisher information, the variance of $\log{(\rho)}$ can be derived as
$\varepsilon_{i}=\sqrt{\frac{4\cdot\hat{k}_{i}+2} {\left(\hat{k}_{i}-1\right)\cdot\hat{k}_{i}}}$.
In Supp. Inf. Text S1 we provide the derivation of this expression and a pseudocode for obtaining the PA$k$ estimation. More details and a validations of the procedure can be found in ref.~\cite{FEwithoutCV}.

\subsubsection{\label{sec:Peaks} Automatic detection of density peaks}

The first step of our algorithm is finding automatically the density peaks.
As in the standard Density Peaks (DP) clustering, we here assume that the density peaks are surrounded
by neighbors with lower local density and {that they are} at a relatively large distance
from any points with a higher local density. However, while in the standard DP algorithm this definition is left to the interpretation of the so-called decision graph, in this section we  show that the additional information provided by PA$k$ (an estimate of the error in the density and the neighborhood size around each point in which the density can be considered approximately constant) can be exploited to provide a quantitative definition of density peaks, therefore allowing their {unsupervised} detection.

To {detect} density peaks we do not directly consider the density of points, that typically varies by several orders of magnitude, but  the logarithm of their density $\log(\rho_i)=-F_i$ identified as the free energy at point $i$ from the PA$k$ estimator~\cite{FEwithoutCV}. 
{As it is}, the estimate of $F_i$ is affected by non-uniform errors, and defining cluster centers as the maxima of the density
as implicitly done in in the DP clustering~\cite{AlexAle} is not always effective. 
We indeed verified that a straightforward combination of DP clustering 
with the PA$k$ density estimator fails to properly localize the correct clusters when the error is highly non-uniform: 
a point whose estimated $\log(\rho_i)$
is large but is affected by a comparatively large error $\varepsilon_i$ is not likely to be a genuine density peak.
To tackle this problem we developed {a} heuristic approach 
for defining cluster centers, {which provides reliable results in practical applications, as we show in the following}. We define as cluster centers the local maxima of $g_{i}$, where $g_i$ is defined as
\begin{equation}
        g_{i}=\log(\rho_i)-\varepsilon_i.
    \label{eq:gi}
\end{equation}
This definition is a generalization of the one used in ref.~\cite{AlexAle}, since the local maxima of $g_{i}$ coincide with the local maxima of $\rho_{i}$ if the error is uniform. If the error is not uniform, points with large error are less likely to be selected as local maxima with respect to points with a small error.
Following ref.~\cite{AlexAle}, we then compute $\delta_{i}={\displaystyle \min_{j:g_{j}>g_{i}}r_{ij}}$, namely the distance to the nearest point with higher $g$,
and we automatically find the cluster centers using the following heuristic.
\begin{HR1}
        \label{hr1}
        \begin{minipage}[t]{\linegoal}
        {{We consider point $i$ a putative center if it satisfies the following property: all its $\hat{k}$ nearest neighbors contributing to determine the value 
        of its density have a value of $g$ lower than $g_{i}$}. As a second condition a center can not belong to the neighborhood of any other point with higher $g$:
        \begin{enumerate}[font=\normalfont,leftmargin=*]
            \item $\delta_{i}>r_{\hat{k}_{i}}$
            \item $\forall\ j$ with $g_j > g_i$ : $i \notin NN_j := \{k: r_{jk}<r_{\hat{k}_{j}}\}$ 
        \end{enumerate}}
        \end{minipage}
\end{HR1}

In this heuristic, $\hat{k}_{i}$ is the optimal number of nearest neighbors defined as in ref.~\cite{FEwithoutCV}, and $NN_j$ denotes the optimal neighborhood of $j$.
Moreover, the second criterion in Heuristic~\ref{hr1} 
makes putative centers selection more robust in front of statistical fluctuations in the neighborhood estimation, see Figure~\ref{fig:heuristic} A for an illustrative example. A pseudocode implementing this heuristic is presented in Algorithm 1.
\begin{algorithm}
        \caption{Automatic detection of cluster centers}
        \label{alg1}
        \begin{algorithmic}[1]
                \Require {Set of points $N_{\textrm{points}}\  x_1,\dots,x_N \in \mathbb{R}^{d}$}
                \Ensure  {\emph{Centers}, the list of points identified as centers.}
                \State $\log(\rho)$, $\epsilon$, $\hat{k}$ $\leftarrow$ PA$k$($N_{\textrm{points}}$) \Comment{where $\rho$-density, $\epsilon$-error, $\hat{k}$-optimal neighborhood size}
                \State $g := \log(\rho)-\epsilon$ \Comment{compute $g$ for all points in $N_{\textrm{points}}$}
                \State \emph{Centers} $\leftarrow$ empty array
                \For {$i \leftarrow 1, N$} \Comment{(criterion~1 in Heuristic~\ref{hr1})}
                    \State \emph{putative\_center} $\leftarrow$ True
                    \For {$j \leftarrow 1, \hat{k}_i$} \Comment{loop over the points in $NN_i$, the neighborhood of $i$}
                        \If {$g_j > g_i$}
                            \State \emph{putative\_center} $\leftarrow$ False
                            \State \textbf{break}
                        \EndIf
                    \EndFor
                    \If {\emph{putative\_center}=True} \Comment{the closest point with higher density is not in $NN_i$, thus Criterion~1 is satisfied.}
                        \State add $i$ to \emph{Centers}
                    \EndIf
                \EndFor

                \For {$i$ in \emph{Centers}}  \Comment(criterion~2 in Heuristic~\ref{hr1})
                    \For {$j \leftarrow 1, N$}
                        \If {$g_j>g_i$ and $j \in NN_i$}
                            \State remove $i$ from \emph{Centers} \Comment{$i$ belongs to the neighborhood of another point with higher $g$}
                        \EndIf
                     \EndFor
                 \EndFor
        \end{algorithmic}
\end{algorithm}

In Figure~\ref{fig:Clusterexpl}B we show the decision graph (i.e., the value of $\delta_{i}$ as a function of $g_i$) for a sample of 20000 points
extracted from the probability density distribution shown in Panel~A.
The points surrounded by a circle are those that are automatically chosen as
putative centers according to Heuristic~\ref{hr1}.

The next step is to assign all the points that are not centers to the same cluster as the 
nearest point with higher $g$. This assignation is performed in order of decreasing $g$.
{Choosing the points highlighted in Figure~\ref{fig:Clusterexpl}B as centers} leads to
a high splitting of the data set (see  Figure S1 for the result of this preliminary
assignation). Indeed, Heuristic~\ref{hr1} correctly identifies the genuine
probability peaks but also the spurious statistical fluctuations of the density induced
by the finite sampling. Consequently, we developed a protocol to assess the peaks significance,
which allows distinguishing meaningful density peaks from statistical
fluctuations of the density, as explained in the following sections.

\subsubsection{\label{sec:Saddles}Finding the saddle points}

We here introduce a procedure that allows finding the saddle points {of the
probability density function} between each peak and its neighboring {ones, which will be crucial for providing insights on the structure of the data.}
We first find the data points that are at the border
between two clusters following Heuristic~\ref{hr2}.
\begin{HR1}
        \label{hr2}
        \begin{minipage}[t]{\linegoal}
        {A point $i$ belonging to cluster $c$ is assumed to be at the border between cluster $c$ and $c'$ if its closest point $j$ belonging to $c'$ is within a distance $r_{\hat{k}_{i}}$ and if $i$ is the closest point to $j$ among those belonging to $c$:
        \begin{enumerate}[font=\normalfont,leftmargin=*]
                \item Let $j =\arg{\displaystyle \min_{k: k \in c'}{r_{ik}}}$, $r_{ij}<r_{\hat{k}_{i}}$  
                \item  $i =\arg{\displaystyle \min_{k: k \in c}{r_{jk}}}$ 

        \end{enumerate}}
        \end{minipage}
\end{HR1}
\begin{figure}[ht!]
\begin{center}
\centerline{\includegraphics[width=0.84\linewidth]{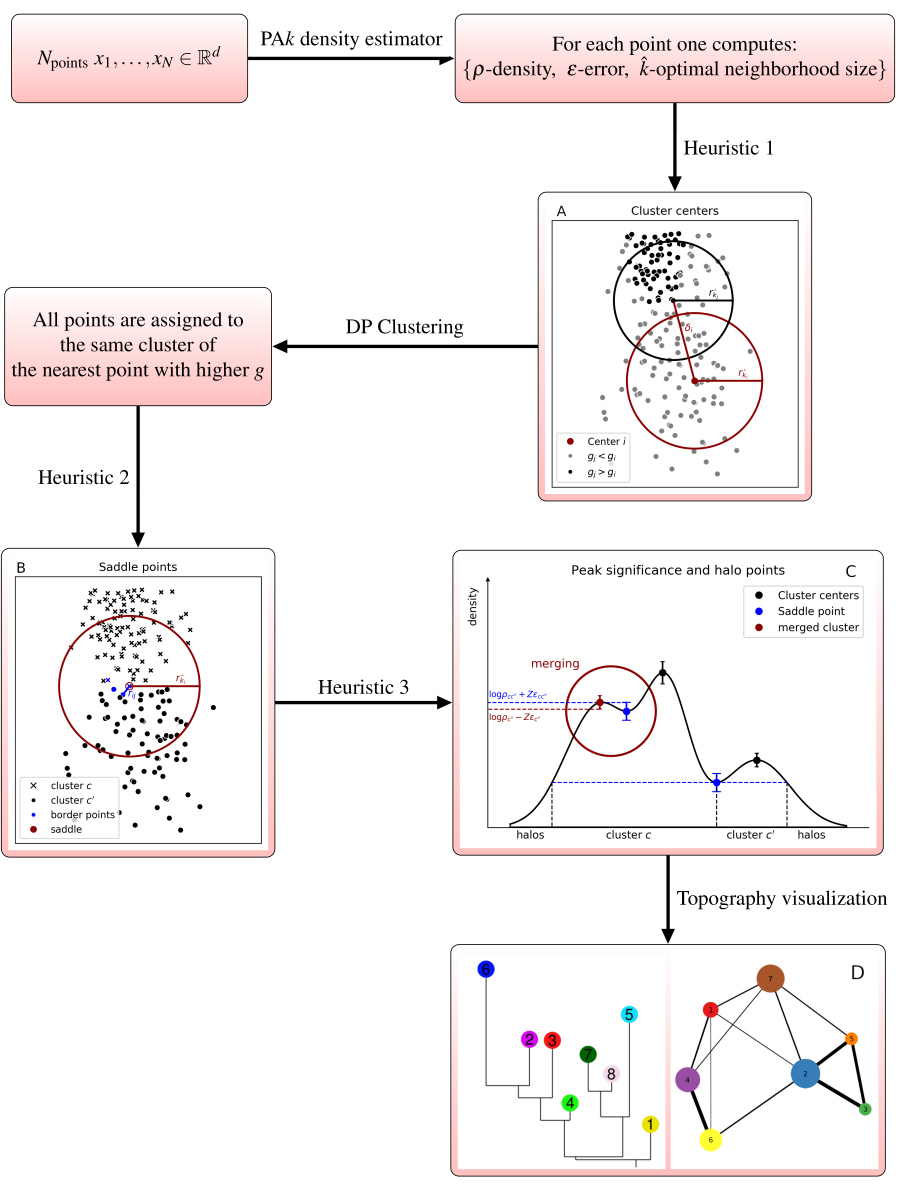}}
        \caption{Flowchart representation of the  algorithm described in this work. The novel procedure are illustrated with two dimensional
        examples: (A) Automatic detection of cluster centers (Heuristic~\ref{hr1}), (B) Saddle points between cluster $c$ and $c'$
        (Heuristic~\ref{hr2}), (C) Merging of the clusters induced by statistical fluctuation (Heuristic~\ref{hr3} and (D) Topography
        representation }\label{fig:heuristic}
\end{center}
\end{figure}

The saddle point {of the probability density function} between a pair of clusters $c$ and $c'$ is defined as the point with the highest value of $g$ among those at the border between $c$ and $c'$, see Figure~\ref{fig:heuristic}B for an illustrative example.
The value of the logarithm of the density of this point and its error are denoted by
$\log{\rho_{cc'}}$ and $\varepsilon_{cc'}$.
The border points between the clusters reconstructed from the two dimensional example in Figure~\ref{fig:Clusterexpl} are shown in black in Panel~C, while the saddle points are circled in blue. The pseudocode implementing this heuristic is presented in Algorithm 2.

\begin{algorithm}
        \caption{Saddle points between clusters $c$ and $c'$}
        \label{alg2}
        \begin{algorithmic}[1]
                \Require {Neighborhood size $\hat{k}$ and the cluster label assigned to each point in $N_{\textrm{points}}$}
                \Ensure  {\emph{Border}$_{cc'}$, the list of points at the border between $c$ and $c'$.}
                \State \emph{Border}$_{cc'}$ $\leftarrow$ empty array
                \For {$i$ in $c$} \Comment{(criterion~1 in Heuristic~\ref{hr2})}
                    \For {$j \leftarrow 1, \hat{k}_i$} \Comment{loop over the points in $NN_i$, the neighborhood of $i$}
                        \If {$j$ in $c'$}
                            \State \emph{candidate} $\leftarrow$ $j$
                            \State \textbf{break}
                        \EndIf
                    \EndFor

                    \For {$k \leftarrow 1, \hat{k}_j$}  \Comment(criterion~2 in Heuristic~\ref{hr2})
                        \If {$k = i$}
                            \State \emph{Border}$_{cc'}$ $\leftarrow$ \emph{candidate}
                            \State \textbf{break}
                        \EndIf
                        \If {$k$ in $c$}:
                            \State \textbf{break}  \Comment{$i$ is not the closest to $j$}
                        \EndIf
                    \EndFor
                 \EndFor
        \end{algorithmic}
\end{algorithm}

\subsubsection{\label{sec:Merging}Assessing the peaks and assignation significance}

Based on the value of $\log{(\rho_{cc'})}$, the {logarithm of the density at the saddles}, and their error we
introduce a criterion for distinguishing genuine density peaks from statistical
fluctuations of the density due to finite sampling, as defined by Heuristic~\ref{hr3}.
\begin{HR1}
        \label{hr3}
        \begin{minipage}[t]{\linegoal}
        {A cluster $c$ is considered as the result of a statistical fluctuation if all the points assigned to it
                have density values compatible, within their errors, with the border density. The cluster is thus merged
                with a neighboring cluster \mbox{$c'$} if:
        \begin{enumerate}[font=\normalfont,leftmargin=*]
                \item $\left(\log{\rho_{c}}-\log{\rho_{cc'}}\right)< Z\cdot\left(\varepsilon_{{c}}+\varepsilon_{{cc'}})\right.$
        \end{enumerate}
                where $\rho_{c}$ is the density of the center of cluster $c$.}
        \end{minipage}
\end{HR1}

The constant $Z$ entering Heuristic~\ref{hr3} fixes the level of statistical confidence at which one decides to consider a cluster meaningful. It is the only  free parameter
of our approach, but its value has a clear statistical interpretation (see Figure~\ref{fig:heuristic}C for an illustrative example).
Heuristic~\ref{hr3} is checked for all the clusters $c$ and $c'$ in order of
decreasing $\log{\rho_{cc'}}$.
Therefore, the implemented procedure performs a multimodality test
based in the error estimated with PA$k$, which
iteratively compares the difference between the estimated log-density at each pair of
peaks (clusters) and the log-density estimated at the saddle point of the
probability density function between them.
This procedure prunes the set of clusters from those
corresponding to density maxima that are not statistically robust, thus recovering the
topography of the underlying probability function.

Furthermore, the knowledge of the border densities between clusters allows to identify the set of points whose assignation is not reliable~\cite{AlexAle}. Indeed, DPA inherits from standard DP clustering the capability of classifying the points as `core' (those whose assignation is robust) or `halo' (they are assigned to a cluster but with a lower level of confidence). Points are classified as `halo' if their density is lower than the highest border density between the cluster at which they are assigned and any other cluster. 

The pseudocode implementing this heuristic is presented in Algorithm 3.

\begin{algorithm}
        \caption{Assessing peaks significance}
        \label{alg3}
        \begin{algorithmic}[1]
                \Require  {\emph{Border}$_{cc'}$, the list of points at the border between $c$ and $c'$.}
                \Ensure  {\emph{Topography}: $N_{clus}\times N_{clus}$ symmetic matrix, in which the diagonal entries are the height of the
                peaks and the off-diagonal entries are the heights of the saddle points.
                          \emph{clu\_lables}: the final assignation to clusters after merging.
                          \emph{halos}: points with not reliable assignation.}
                \State \emph{apply\_merging} $\leftarrow$ True
                \While {\emph{apply\_merging} is True}
                    \State \emph{apply\_merging} $\leftarrow$ False
                    \For {points in \emph{Border}$_{cc'}$}
                        \If {Condition (\ref{hr3}) is True}  \Comment{(criterion~1 in Heuristic~\ref{hr3})}
                            \State \emph{apply merging} $\leftarrow$ $True$
                        \EndIf
                    \EndFor
                    \If {\emph{apply\_merging} is True}
                        \State {$(c_{h},c_{l})$} $\leftarrow$ pair with the highest $\rho_{c,c'}$ \Comment{$c_{h}$ is the peak with highest $\rho$}
                        \State {merge $c_{l}$ with $c_h$}
                        \State {update \emph{Topography}}
                        \State {update \emph{clu\_labels}}
                    \EndIf
                \EndWhile

                \State \emph{halos} $\leftarrow$ empty array \Comment{Find halo points}
                \For {$i$ in $N$} \Comment{loop over the point in the data ser}
                    \If {$\rho_i <$ min\{$\rho_{c,c'}$\}}
                        Add $i$ to \emph{halos}
                    \EndIf
                \EndFor
        \end{algorithmic}
\end{algorithm}

In Figure~\ref{fig:Clusterexpl}D it can
be seen that the cluster assignation after the merging (with $Z=1.5$) resembles almost
perfectly the peaks shown in Panel~A (black points correspond to halo).
Indeed, these results {will} correspond with those obtained by the standard Density Peaks
method if by visual inspection one chooses as centers the colored circles in Panel~B.

\subsubsection{\label{sec:Topo}Representing the topography}

The information about the location and the height of the saddles allows building a compact
representation of the topography of the probability distribution function from which the 
data points are harvested.
To visualize the topography of the density distribution we follow two paths. 
Both are based on the fact that the higher the density of the border
between clusters, { the more similar such clusters} can be considered. Therefore,
we define the distance between two clusters as follows:
$d_{cc^{'}}=\max(\log{(\rho_i)})-\log{(\rho_{cc'})}$.
One possible way to visualize the topography is constructing a hierarchical tree
by applying the Single Linkage algorithm~\cite{SingleLinkage} using $d_{cc^{'}}$ as distance.
This representation is similar to the one used in hierarchical
density based methods~\cite{hartigan1981consistency,HDBSCAN,RobustSL,RobustDB}.
In our case, to encode more information when representing the tree, the height of the
branches is proportional to the density of the peak associated to them and the separation
between branches in the x-axis is proportional to the population of the clusters. An example 
of this representation is provided in  Figure~\ref{fig:Clusterexpl}E.

An alternative way {to represent the topography is by projecting the clusters in two
dimensions} and visualizing their relationship as a network where the thickness of the links 
between clusters is proportional to the log-density at the border. This {leads} to a 
representation similar to those used in Markov State Model analysis~\cite{MSM2010}. An 
example of this representation is provided in Figure~\ref{fig:Clusterexpl}F. To encode more 
information in a single plot, the area of the circles representing the clusters is 
proportional to their population.

Both {visualizations} provide complementary information about the underlying probability density
function. In the example shown in Figure~\ref{fig:Clusterexpl} the hierarchical relationship
between clusters 2, 3 and 6 (magenta, red and blue) is more evident in the tree 
representation. However, the close contact between clusters 4 and 8 (light green and light 
pink) is evident only using the network representation.
Additional examples are shown in Supp. Inf. (Figures S2-5).
\subsubsection{\label{sec:signif}Statistical significance of the clusters}

The value of $Z$ in the Heuristic~\ref{hr3} is used to control the statistical reliability of
the density peaks.

In general, at low $Z$ values the method is more sensitive to variations of the density,
but fluctuations due to sampling artifacts are also {identified} as clusters density. Then, the
higher the value of $Z$, the lower the sensitivity to density changes, but the higher
the statistical reliability of the peaks.
If the sampling of the probability distribution function is good enough, one can
increase the value of $Z$ in order to enhance the statistical confidence. If the sampling
is poor (something that easily happens if the intrinsic dimension of the data is high), one
is forced to accept a lower level of confidence (lower value of $Z$) and a significant probability of observing
some spurious clusters.  
\begin{figure}[ht!]
                \centerline{\includegraphics[width=1.0\linewidth]{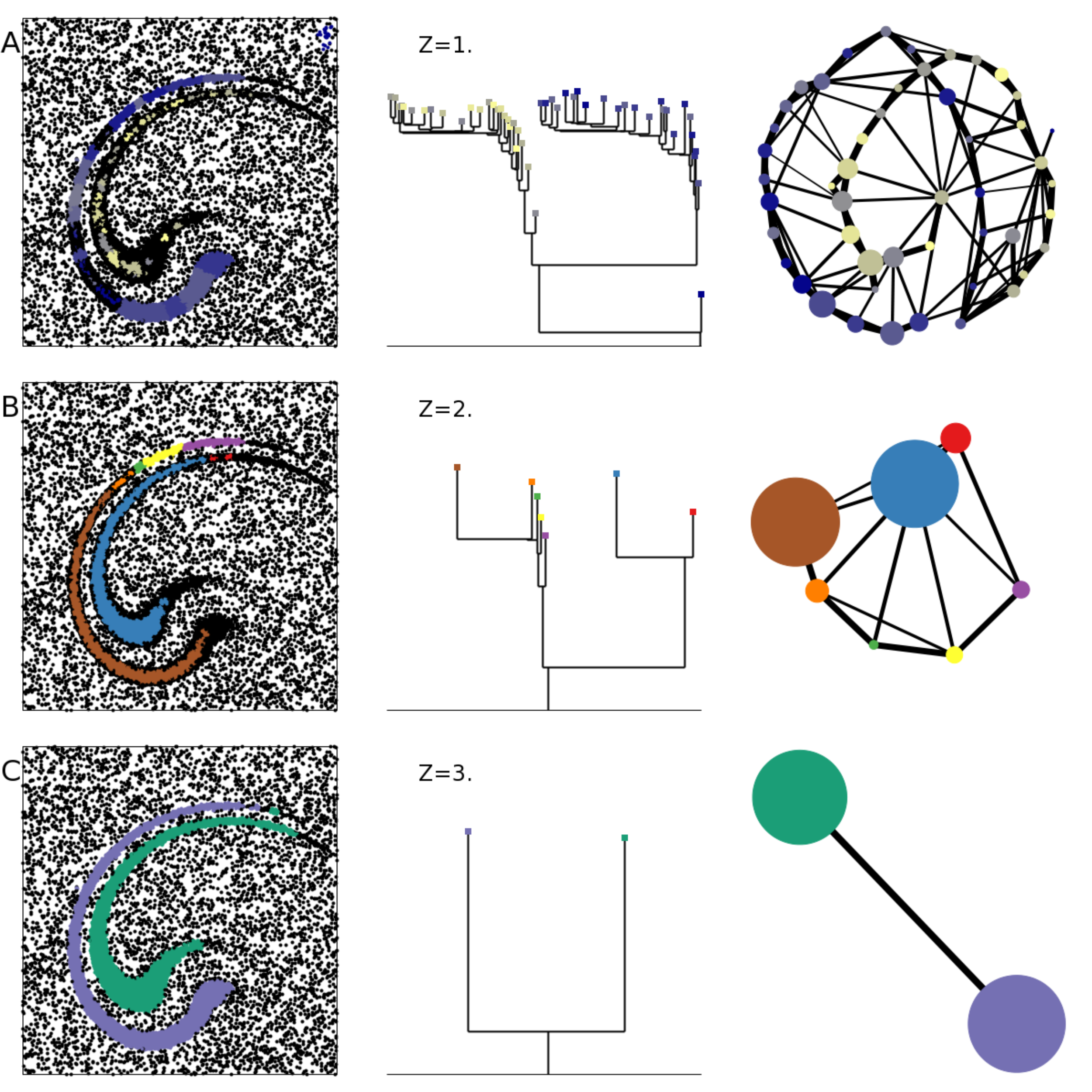}}
			\caption{Results of the topography reconstruction on SPIR2, a data set with two spirals, at different values of $Z$ (Panel~A, $Z=1.$, Panel~B $Z=2.$ and
                Panel~C $Z=3.$. The colors in the dendrogram (center) and in the network
                (right) correspond to those in the assignation (left).}\label{fig:spirali}
\end{figure}
In Figure~\ref{fig:spirali}  we show an example of how the
parameter $Z$ affects the clustering classification in a well sampled distribution. Although
the best results are obtained with $Z=3$, and the number of clusters increases for smaller $Z$,
the topography of the data set allows identifying the two main peaks of the
distribution at any value of the parameter $Z$.
A systematic study of the role of this parameter {and its impact} in detecting spurious clusters is provided in Supp. Inf. Text~S1 and Figure~S6.

\section{\label{sec:Results}Results and Discussion}
 \subsection{\label{sec:valid} Validation on artificial data sets}

We first validate our approach on artificial two-dimensional data sets where the ground-truth classification is available. We perform benchmarks on the sets represented in Figure~\ref{fig:Clusterexpl} (hereafter called CLUS8), in \mbox{Figure~\ref{fig:spirali}} (SPIR2), and in Figure S2 (AGGR), S3 (SPIR3) and S4 (HORSE) in Supp. Inf., which have a number of clusters ranging between two and eight. 
In all the cases the ground-truth classifications are obtained using the explicit probability density function from which the points are generated:
the ground-truth clusters correspond to maxima in this function.
In detail, for each maximum we find the region such that: (i) contains the maximum; (ii) its boundary is a contour level of the probability density function passing through a saddle point; (iii) no other saddle points, except the one on the boundary, are contained in it.  Data points belonging to this region form a ground-truth cluster, while the points which do not belong to any of these regions are classified as `noise'. 
The artificial data sets and the corresponding ground-truth classifications are provided as supplementary material.

To assess the similarity between the DPA clustering assignation and the ground-truth classifications we use the Normalized Mutual Information (NMI)~\cite{AMI}. Note that when the two sets of labels have a perfect one-to-one correspondence, the NMI is equal to one. Since in the ground-truth some points are classified as `noise', we quantify the accuracy in two different ways. (i) We check the accuracy in the clustering assignation of the points that {belong} to the ground-truth density peaks: we classify all the points using the DPA algorithm without labelling any point as `halo', and compute the NMI by excluding the `noise' points, which is denoted in the following as NMI$_{dpa}$. (ii) We check the accuracy of all points assignation, including those classified as `noise': by following the protocol described in Methods~\ref{sec:Merging} we find the `halo' points that should ideally correspond to the `noise' classification, and then compute the NMI for all the points considering `noise' as a ground-truth label, which is denoted in the following as NMI$_{halo}$.
We also provide the False Negative and False Positive ratios for noise detection using halo points as noise classification (FNR$_{halo}$ and FPR$_{halo}$ respectively).

For the data sets CLUS8, SPIR2, AGGR, SPIR3 and HORSE we find the interval values of $Z$, the only relevant parameter in our approach, maximizing NMI$_{halo}$. This interval is reported in Table~\ref{tab:NMI} as $[Z_{best}]$, together with the corresponding value of NMI$_{halo}$. 
In the same table we also report the interval  $[Z_{90\%}]$ within which the  NMI$_{halo}$ is at least $90\%$ of its maximum value, and $N_{clus}$ as the number of clusters inferred for $Z$ in $[Z_{best}]$.

\begin{table*}[ht!]
\caption{Validation of the DPA clustering on artificial data sets. We use a selection of two-dimensional benchmark data sets having a number of clusters ranging between two and eight: data set CLUS8 represented in Figure~\ref{fig:Clusterexpl}, SPIR2 in \mbox{Figure~\ref{fig:spirali}}, and AGGR, SPIR3, and HORSE represented in Supp. Inf. in Figure S2, S3 and S4 respectively. To assess the similarity between the DPA clustering assignation and the ground-truth classifications we measure the Normalized Mutual Information~\cite{AMI}(NMI) in two different ways: NMI$_{dpa}$ and NMI$_{halo}$.  NMI$_{dpa}$ is used to assess the accuracy in the clustering assignation of the points that belongs to the ground-truth density peaks, thus excluding the `noise' points. NMI$_{halo}$ is used to assess the accuracy of all points' assignation, by including those labelled as `halo' (described in Methods~\ref{sec:Merging}) that should ideally correspond to the `noise' classification. For each data set in the table we report $[Z_{best}]$ as the interval values of $Z$ that maximize the NMI$_{halo}$, together with the corresponding value of NMI$_{halo}$. We report the interval  $[Z_{90\%}]$ within which the  NMI$_{halo}$ is at least $90\%$ of its maximum value, and $N_{clus}$ as the number of clusters inferred for $Z$ in $[Z_{best}]$. We also provide the False Negative and False Positive ratios for noise detection using halo points as noise classification (FNR$_{halo}$ and FPR$_{halo}$ respectively).
}
\centering
\begin{tabular}{|c||c|c|c||c||c|c|c|}
\hline
\cline{2-7}
 & $[Z_{best}]$ & $[Z_{90\%}]$ & N$_{clus}$ & NMI$_{dpa}$ & NMI$_{halo}$ & FNR$_{halo}$ & FPR$_{halo}$ \tabularnewline
\hline
\hline
CLUS8 & $[1.5,1.9]$ & $[1.1,4.0]$ & 8 & 0.996 & 0.841 & 0.010 & 0.111\tabularnewline
\hline
SPIR2 & $[2.9,4.0]$ & $[2.9,4.0]$  & 2 & 0.966 & 0.929 & 0.002 & 0.015\tabularnewline
\hline
AGGR & $[2.0,2.5]$ & $[1.6,2.8]$ & 7 & 0.994 & 0.844 & 0.070 & 0.053\tabularnewline
\hline
SPIR3 & $[2.5,3.6]$ & $[2.5,3.6]$ & 3 & 0.996 & 0.809 & 0.000 & 0.100\tabularnewline
\hline
HORSE & $[2.6,4.0]$ & $[2.6,4.0]$ & 3 & 0.987 & 0.832 & 0.077 & 0.036\tabularnewline
\hline
\end{tabular}

\label{tab:NMI}
\end{table*}

If the value of $Z$ is appropriately chosen, the clustering algorithm is able to find all the ground-truth density peaks in all the considered artificial data sets with values of NMI$_{dpa}$ close to $0.99$, as shown in Table~\ref{tab:NMI}.
However, the optimal choice for $Z$ is not an isolated value, and the performance of the DPA clustering is not critically dependent on its exact choice: $[Z_{90\%}]$ show that already excellent results are obtained for a significantly large interval of $Z$ values.
The NMI estimated on the full set of points (NMI$_{halo}$) is also very large for most data sets. The worst performance is observed in SPIR3, where  NMI$_{halo}$ is equal to $0.809$. 
This can be explained by a FPR$_{halo}$ of 0.1 corresponding to a misclassification as `halo' of $10\%$ of data points that should instead belong to a cluster according to the ground-truth. A relatively large value of FPR$_{halo}$ is observed also in the data set CLUS8. Indeed, the `halo' by definition identifies all points whose assignment to a cluster is not reliable: this includes the `noise' points, but also other points in low density regions. 
The values for FNR are instead very small, and even zero at times, indicating that the approach provides a reliable classification of the `halo' points

 \subsection{\label{sec:MNIST}Clustering a handwritten digits data set}
\begin{figure*}
         \centering{\includegraphics[width=0.54\linewidth]{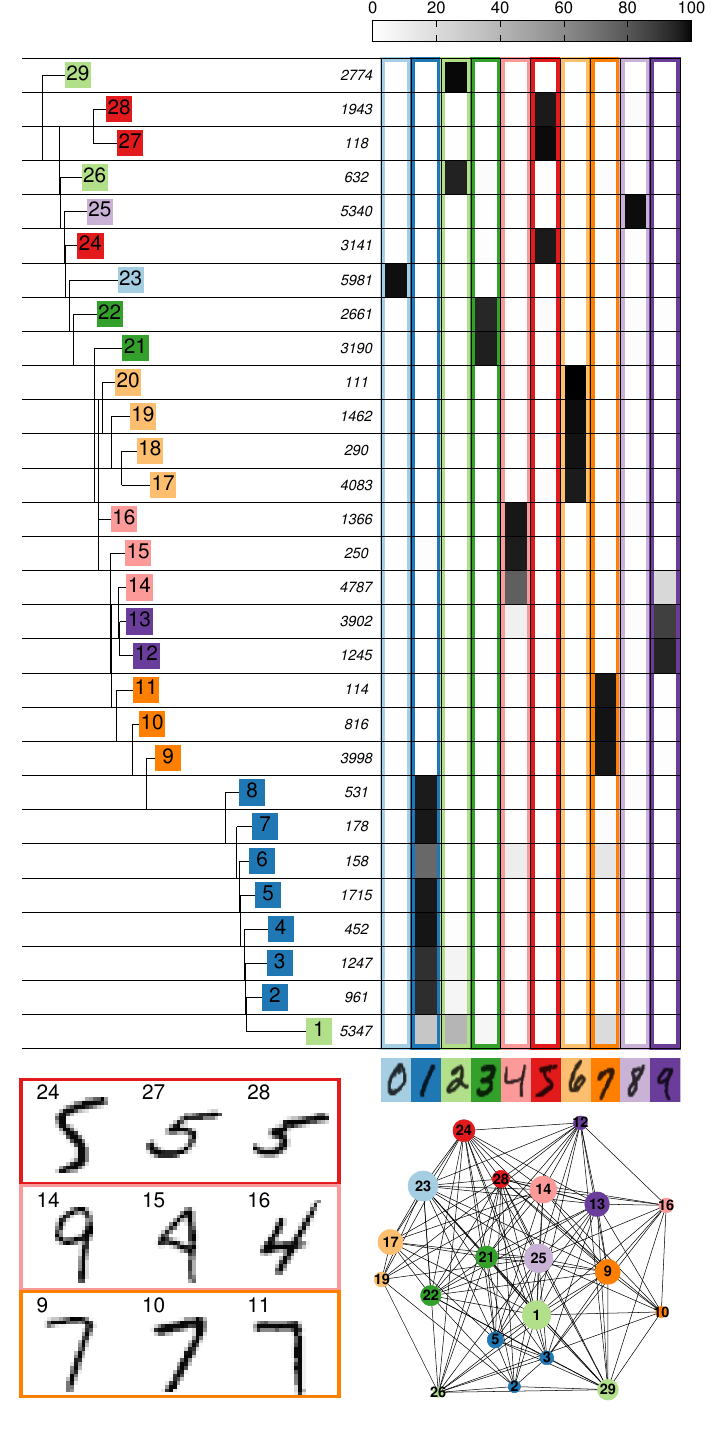}}
         \protect\protect\caption{Cluster analysis of the MNIST data set. In the matrix we
     represent, for clusters with a population greater than 100, the fraction of
     elements assigned to a cluster belonging to each of the ground truth labels. The
     higher the fraction, the darker the cell. On the left we show {the clusters dendrogram}. In correspondence to each leaf we indicate the cluster label and its
     population. To simplify the interpretation, {the clusters background color} is chosen according with a majority rule. For visualization purposes, only
     clusters with population higher or equal than 100 are shown. Bottom right: the clusters
     with population higher than 600 in a network representation. Bottom left: images
     corresponding to the centers of clusters assigned to numbers $5$, $4$ and $7$.
     }
     \label{fig:MNIST}
     \end{figure*}

We further test our approach on the MNIST~\cite{MNIST} data set, which includes 60000 images
of handwritten digits between 0 and 9. We compute the pairwise distances using the tangent
distance~\cite{TanDist}, a metric explicitly developed for image comparison 
that is less sensible to transformations like rotation or translation. 
The intrinsic dimension of the data set estimated by the TWO-NN approach~\cite{faccoID} is 8, and the value of $Z$ is set to 1.6. 
The results are summarized in Figure~\ref{fig:MNIST}, {with} the topography description 
represented by the dendrogram in the {top} left panel and by the network at the bottom right. 
The color of each cluster
in the topography representations derives from a majority rule assignation, {namely
the color is the one corresponding to the digit label with higher presence in the cluster
whose color code is indicated under the matrix}.
The number of elements in each cluster is shown in italics between the dendrogram and the
matrix {that represents in a grey palette the fraction of points assigned to a cluster belonging to each of the ground truth labels: the darker the cell, the higher the fraction.} 
As it can be seen {from the matrix}, the
clusters contain almost always data points with  a consistent  ground truth classification. The number of clusters is larger than 
the ten classes of digits in the ground truth. For example, 
number seven is split in three clusters (9 10 and 11) {and} number one in seven clusters. However, those belonging to the same digit appear as closely related to each other in the hierarchical structure represented by the dendrogram.  
The origin of this splitting can be {in many cases} ascribed to real differences in the handwritten digits: the same ground truth label is often assigned to images that look qualitatively different. Some examples are provided in the bottom left panel in  Figure~\ref{fig:MNIST}, {where the}  
images belonging to cluster 24, 27 and 18 (number five), and those belonging to cluster 14, 15 and 16 (number four) look indeed qualitatively different.  
In the same panel we also show representative images of clusters 9, 10 and 11 (number seven) {which look instead  similar}. Therefore, in this case the splitting of the three clusters is likely to be an artifact of our approach.
With the aim of quantitatively measure the quality of the clustering result, we computed the NMI 
between the digit label assigned to each cluster according to a majority rule and the ground truth. We find a NMI of $0.84$.
{A further analysis of this data set was done after undersampling it. In this case, only
10000 out of 60000 images were analyzed.} 
The computed intrinsic dimension with TWO-NN~\cite{faccoID} is equal to 7, {which} is coherent with the loss of information
due to the undersampling, and the parameter $Z$ is {lowered to $1.0$
to achieve a sufficient sensitivity to data structures, although accepting a possibly larger 
number of spurious clusters, as discussed in the Methods.}
The results are very similar to those {obtained by analyzing} the complete data set. The main difference is
that the mixing between the {digits four and nine in the reconstructed clusters} is more significant, leading to a
lower NMI of $0.76$. The confusion matrices used for the NMI calculation for the full and the undersampled data set are respectively provided in Supp. Inf. Table~S1 and S2.

 \subsection{\label{sec:PUA}The density topography of the PUA proteins clan}

We finally exploit the approach introduced in this work to reconstruct
the density topography of a sample of 9684 sequences extracted
from the Pfam clan PUA. The Pfam database~\cite{PFAM} is a large collection
of protein families, grouped into clans or superfamilies; PUA is a
complex superfamily organized into ten families with a population
ranging from a few hundreds to thousands proteins, with many families 
containing a variety of protein architectures.

We first compute the local pairwise distances between the sequences by
using a Modified Hamming distance described in ref.~\cite{IDprot}.
The intrinsic dimension of the data set estimated by TWO-NN~\cite{faccoID} is
equal to 9.
If one uses the standard $k$-NN density estimator to cluster the PUA protein
sequences the choice of the optimal global $k$ is far from trivial.
The PA$k$ density estimator, not surprisingly, finds a huge variability
in the optimal values for $k$, ranging from 3 to 170 across the
data set, reflecting the complexity of the sample. 
{Using the PA$k$ density estimator and $Z$ equal to $2$, we find 123 clusters.}

\begin{figure*}[htp!]
     \centering{\includegraphics[width=0.83\textwidth]{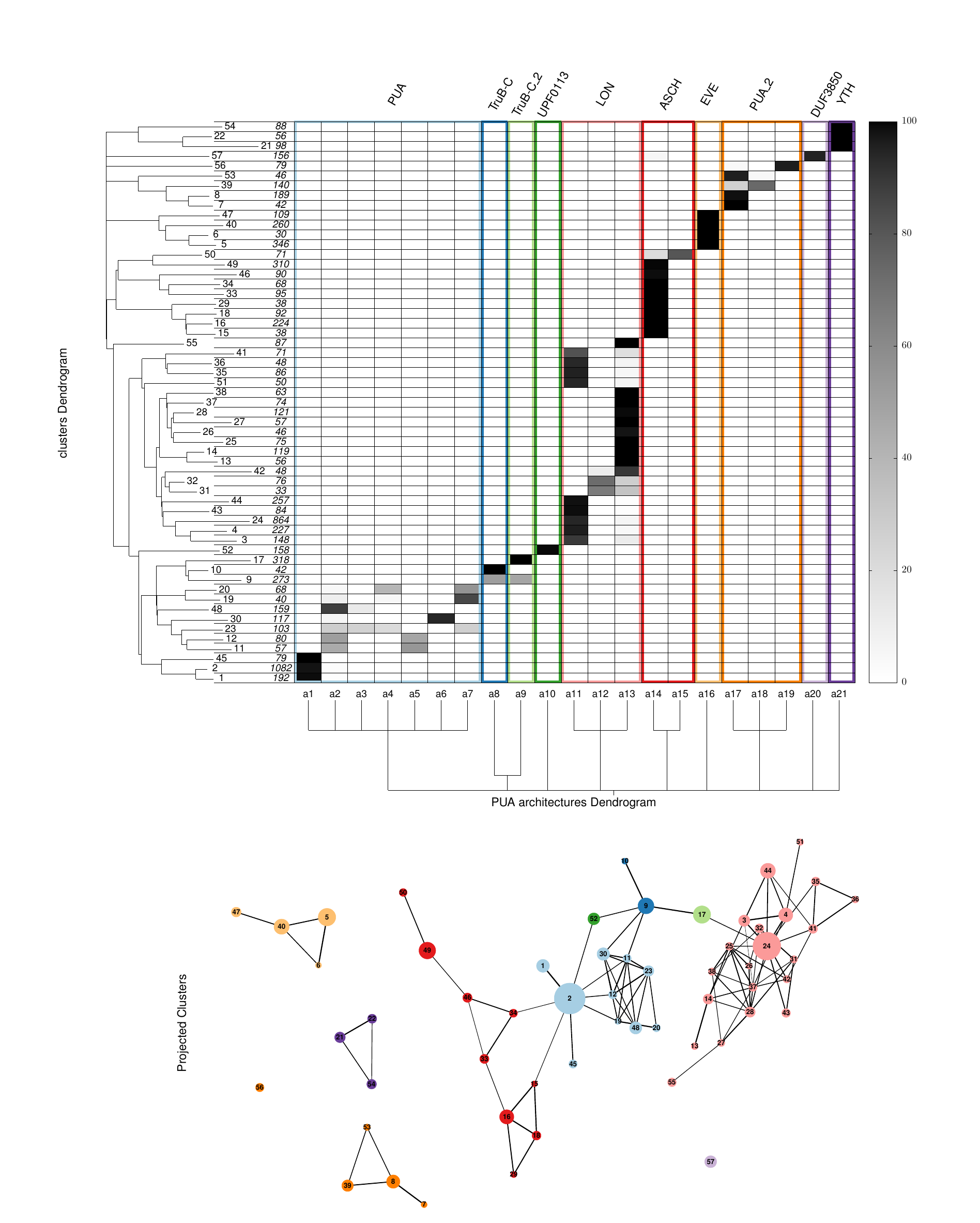}}
     \protect\protect\caption{Cluster analysis of the clan PUA from the Pfam database. We
     represent the Purity Matrix for clusters and architectures with a population greater
     than 40. Color boxes  correspond to families. The purity of the clusters with respect
     to architectures is associated to a grey palette: the darker the cell, the higher the
     purity. On the left of the Purity Matrix we show the dendrogram of the clusters.
     In correspondence to each leaf we indicate the cluster label and the population of the
     cluster. The dendrogram at the bottom is a schematic visualization of the hierarchical
     relationship existing between architectures according to Pfam: architectures connected
     at a higher level (e.g. a1, a2, a3) belong to the same family, while those connected at
     a lower level (e.g. a12, a13) belong to a clan. The Pfam denomination of the
     architectures is provided in Supp. Inf. Bottom: network representation of the clusters represented 
     in the top panel.\label{fig:PUA}}
 \end{figure*}

We test our results against the Pfam classification of PUA sequences into
families, and going into greater detail, into architectures, by computing
the purity of our clusters. Here the purity of a cluster C with respect
to an architecture A is defined as the number of sequences in
C belonging to A divided by the total population of C. In Figure~\ref{fig:PUA}
we represent the correspondence between clusters ordered according
to the dendrogram (y-axis) and architectures (x-axis). Again the network
representation is shown at the bottom of the graph. Only architectures and clusters
with a population greater than 40 are displayed. The Pfam denomination of the
architectures considered in Figure~\ref{fig:PUA} is provided in Supp. Inf.
In this representation the purity of clusters with respect to architectures is
associated to a grey palette: the darker the cell, the higher the purity.
Figure~\ref{fig:PUA} shows that clusters are substantially pure with respect to
architectures (most of the clusters are over 90\% pure).
The quality of the results was also assessed by computing the NMI~\cite{AMI}
of the clustering partition with respect to the Pfam classification. Due to the
hierarchical nature of the method, to compute the indices a family (or architecture)
label is assigned to each cluster according to a majority rule. We find a NMI of
$0.978$ for the classification in families, and  of  $0.871$ for the classification in
architectures, which reveals a high degree of similarity between the clustering partition and the Pfam classification. The considerations that one should take into account when comparing with other methods (Fig.~\ref{fig:performance}) are the same as in the case of MNIST.
The dendrogram provides further information on the complex topography
of the data set, showing, for instance, that clusters belonging to the same
architecture are closely related to each other.
It essentially reflects the similarity between families in the clan as well as their
division into architectures.
The only important exception is that cluster 9 is divided between families TruB-C$\_$2
and TruB-C. These two families are characterized by {sequences with a low similarity} 
within the same family, thus the error in the estimated densities is so large that the faint saddle point
that separates the two families is classified by our algorithm as a statistical
fluctuation.

The network representation shows a complex landscape. For instance, while some families
are well isolated others are interconnected through one or several nodes. The families
PUA and LON, {with the main cluster nodes 2 and 24 respectively,} are divided in many clusters but they are densely interconnected between
them. On the contrary, the family ASCH, although connected, appears to be quite sparse.
The centrality of cluster 9 between families TruB-C$\_$2 and TruB-C is in agreement
with the analysis of the dendrogram.
 \subsection{\label{sec:comp}Comparison with other methods }
We compare our approach with other four state-of-the-art methods for clustering multidimensional data sets:
Spectral  Clustering~\cite{ng2002spectral}, HDBSCAN~\cite{HDBSCAN}, Gaussian Mixture Model~\cite{DPGMM}, 
and DP clustering~\cite{AlexAle}.
The comparison is performed using the artificial data sets CLUS8, SPIR2, AGG, SPIR3 and HORSE considered in Table~\ref{tab:NMI}, 
and the data sets MNIST and PUA introduced in the previous Sections.
Data are first clustered by using the Python implementations available for the selected clustering methods: \emph{BayesianGaussianMixture} and \emph{SpectralClustering} implementations from the \mbox{\emph{scikit-learn}} library~\cite{scikit-learn}, the \emph{hdbscan} library~\cite{hdbscan-python}, and our implementation of the DP clustering.
The similarity between the clustering partitions and the ground-truth classification of the data sets are then evaluated using the NMI~\cite{AMI}.
For the PUA and the MNIST data sets, before performing the NMI calculation, we match the clusters to reference populations using the majority rule: all points within each cluster are labeled as the ground-truth class of highest frequency in the cluster. This way we account for those
true substructures not classified by the ground-truth that are indeed reconstructed as different clusters.

In order to compare all the clustering methods on equal footing, we have to consider that Spectral Clustering, Density Peaks and the Gaussian Mixture Model do not attempt to recognize noise data points, while  HDBSCAN assigns always a noise category.
In the ground-truth of the artificial data sets some points are classified as `noise', while no noise is expected in the PUA and MNIST data sets. 
Therefore we apply method specific calculations of the NMI when considering the artificial data sets or the real world ones.
In particular for Spectral Clustering, Density Peaks and the Gaussian Mixture Model we compute the NMI by ignoring the points classified as `noise' when present in the ground-truth. 
To evaluate the HDBSCAN results on PUA and MNIST data sets instead we compute the NMI by ignoring the points labeled as noise by the algorithm.
In the case of DPA, we can compute the NMI by either assigning all points to a cluster, and ignoring the points classified as `noise' when present in the ground-truth as done for Spectral Clustering, Density Peaks and the Gaussian Mixture Model, or considering the points classified as `halo' in analogy to HDBSCAN.

The accuracy of each considered clustering algorithm is evaluated quantifying the influence of the algorithms parameters on the NMI.
This is done by varying a single parameter of an algorithm while keeping the others at their default values.
For the DPA clustering we compute the NMI as a function of the $Z$ parameter.
For Spectral Clustering, the free parameter is the total number of clusters $N_{cl}$.
For the Bayesian Gaussian Mixture Model, we scan the \emph{weight\_concentration\_prior} parameter for the Dirichlet process as prior type. For this method, the clustering of the high-dimensional data sets PUA and MNIST is preceded by a projection in a 2D space obtained using the t-SNE~\cite{tsne} dimensionality reduction {(these projections, that are informative by themselves, are provided in Figure S7 of Supp. Inf.).}
For the HDBSCAN method, we compute the NMI as a function of the  \emph{min\_cluster\_size} parameter, which fixes the minimum size of a cluster, while we set the parameter \emph{min\_samples} equal to it, as default option.
In the standard DP method, we compute the NMI as a function of the number of clusters, with the peaks chosen in order of decreasing $\gamma=\rho\cdot\delta$. Following the original algorithm~\cite{AlexAle}, the $\delta$ is defined as the distance of a data point from its nearest neighbour of higher density, and the density is estimated by the exponential kernel estimator whose width $d_c$ is chosen in such a way that the average number of neighbors is the $2$\% of the total number of points in the data set. 

\begin{figure}
        \centerline{\includegraphics[width=0.685\linewidth]{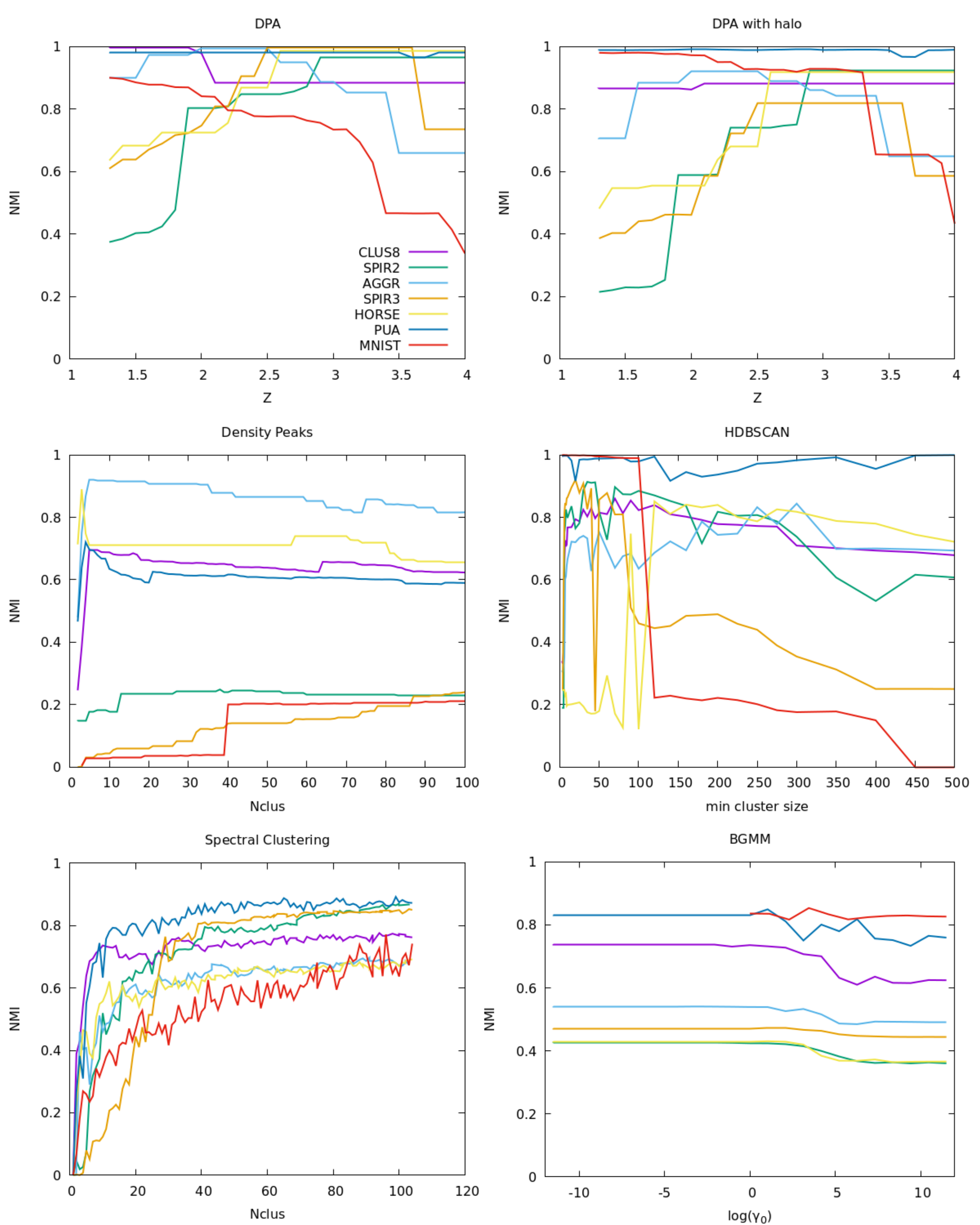}}
        \caption{Performance comparison between the new method, denoted as DPA (Density Peaks \emph{Advanced}), and a selection of clustering methods. The influence of the algorithms parameters on the Normalized Mutual Information (NMI)~\cite{AMI} is evaluated using the artificial data sets CLUS8, SPIR2, AGG, SPIR3 and HORSE considered in in Table~\ref{tab:NMI}, and the real world data sets MNIST and PUA. For Spectral Clustering~\cite{ng2002spectral}, Density Peaks~\cite{AlexAle} and the Gaussian Mixture Model~\cite{DPGMM} we compute the NMI by ignoring the points classified as `noise' when present in the ground-truth. To evaluate the HDBSCAN~\cite{HDBSCAN} results on PUA and MNIST data sets instead we compute the NMI by ignoring the points labeled as `noise' by the algorithm. Panel~A: NMI as a function of $Z$, free parameter for the DPA clustering, computed by assigning all points to a cluster and ignoring those classified as `noise' when present in the ground-truth. Panel~B: NMI as a function of $Z$, free parameter for the DPA clustering, computed by including the `halo' points. Panle~C: NMI as a function of the number of clusters N$_c$, free parameter for DP clustering. Panel~D: NMI as a function of the \emph{min\_cluster\_size}, free parameter for HDBSCAN. Panel~E: NMI as function of the number of clusters N$_{cl}$, free parameter for Spectral Clustering. Panel~F: NMI as function of the \emph{weight\_concetration\_parameter} $\gamma_0$, free parameter for Gaussian Mixture Model.}\label{fig:performance}
\end{figure}

The results are summarized in Fig.~\ref{fig:performance}. 
For the DPA clustering we show the values of NMI computed by assigning all points to a cluster and ignoring those classified as `noise' when present in the ground-truth in Panel~A, and the NMI computed by including the `halo' points in Panel~B. The interval of $Z$ values maximizing the NMI is data set dependent, however by setting $Z=3$ the value of the NMI is larger than $0.8$ for all the data sets in both Panel~A and~B, except for MNIST whose NMI is equal to $0.7$ in Panel~A.
We also notice that in the high-dimensional data sets (MNIST and PUA) the best results are obtained with relatively low values of $Z$: due to the curse of dimensionality the typical error on the density estimate is large, which makes hard to distinguish between real clusters and statistical fluctuations.
In Panel~C we plot the NMI values for the DP clustering. The performance of this algorithm is poor, especially for the data sets MNIST, SPIR2 and SPIR3.  
This is due to the complexity of the density distribution in these data sets, {which makes it difficult} to identify the true clusters centers by selecting the peaks with higher $\gamma$, because the $\delta_i$ parameter is typically very small also for the cluster centers.
In Panel~D we report the value of NMI for HDBSCAN. In data sets SPIR3, HORSE and MNIST, the NMI does not vary smoothly as a function of the \emph{min\_cluster\_size} parameter.
The best results are obtained by choosing a small value of this parameter, but for all the possible choices there are at least two data sets for which the NMI is significantly smaller than $0.8$.  
In Panel~E we report the value of NMI for the Spectral Clustering approach. In this case the best performance is obtained by setting a high value for the free parameter, the number of clusters $N_c$. The  NMI does not decrease when $N_c$ becomes large  because the new clusters are either small or in the points that are assigned as noise in the ground truth.
The best values of the NMI are in many cases worse than those obtained with DPA, in particular for the datasets HORSE, AGG and CLUS8 the best NMI is lower than $0.8$.  
Finally, in Panel~F we plot the value of NMI for the Gaussian Mixture Model: the NMI is remarkably stable with respect to the free parameter, but the performance is in general worse than for the other methods we considered.

 \subsection{\label{sec:cost}Computational cost}
In order to check the run-time performance and scaling of the DPA clustering algorithm, we generated artificial data sets with a distribution defined by the mixture of two Gaussian and background noise, in dimensions ($d$) 2, 4, and 8, and with the number of data points ($N$) ranging from 2000 to 250000. Therefore, 60 data sets with different combinations of $[N,d]$ were generated in total.
For these tests, the time is defined as the total wall-clock time from start to finish, as measured by the \emph{time()} function of the Python \emph{time} module, and for each pair $[N,d]$ the average time and standard deviation are estimated over ten further realization of the data set. The parameter $Z$ has been set to a reasonable large value ($Z=5$) in order to provide an upper bound estimation of the time: as defined by Heuristic~\ref{hr3}, the number of merging steps increases with the value of $Z$, and therefore also the run time of the algorithm.

\begin{figure*}[htp!]
     \centering{\includegraphics[width=0.85\textwidth]{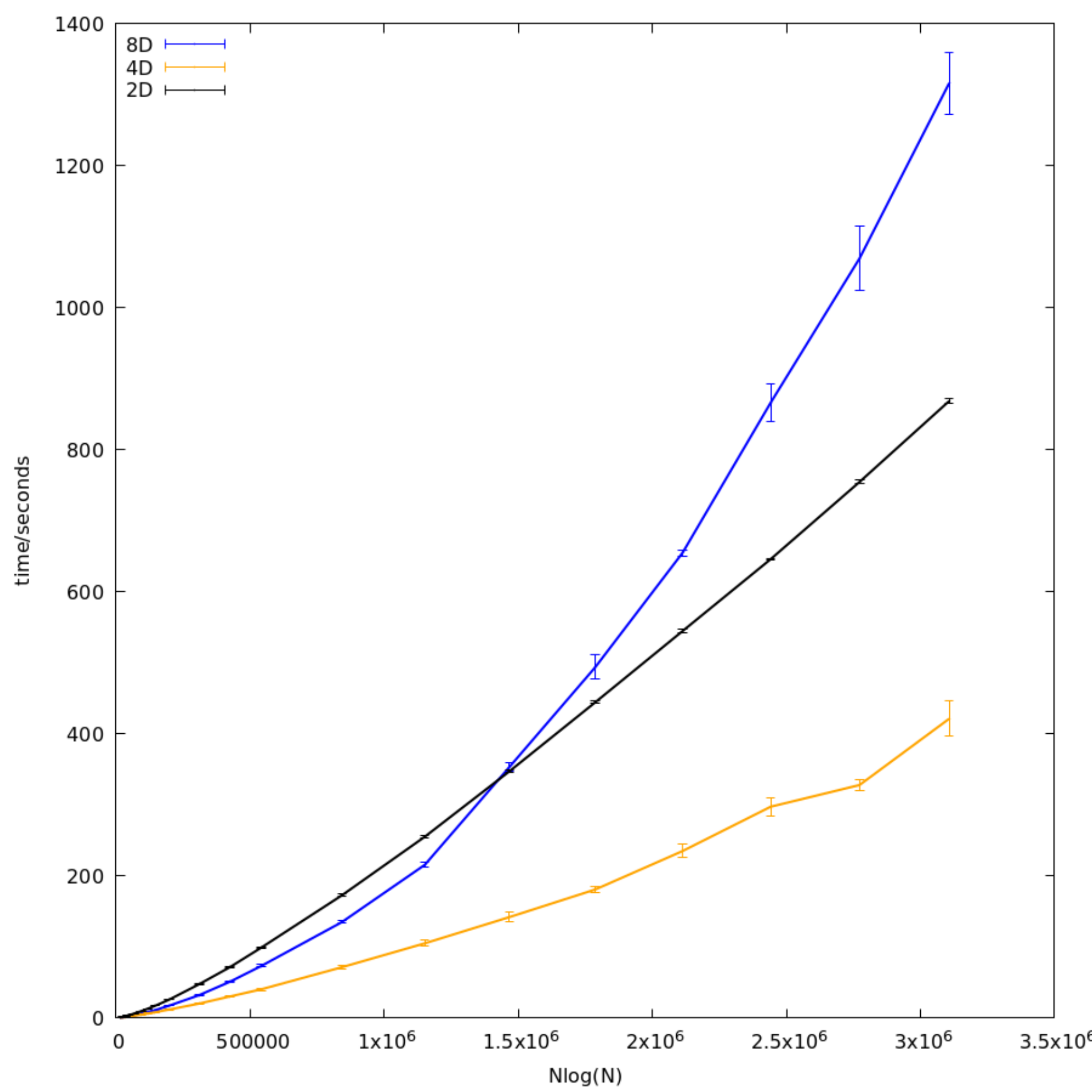}
     \protect\protect\caption{Run times for the current implementation of the DPA algorithm for different number of data points $N$ and embedding dimensions $d$.
     The test cases have been generated from 400000 points that belong to a bimodal Gaussian distribution and then added 100000 points of uniform background noise. Then, they have been undersampled from in a range that goes from 2000 to 250000 points. The procedure has been repeated ten times for obtaining an average time and a standard deviation (error bars). In order to check the scaling with the dimension of the data set, the calculation has been done in dimension 2, 4 and 8. While at low number of samples, the behavior is complex, when they increase it can be seen that the time increases linearly with $N\cdot \log(N)$. The times showed correspond to calculations made in a Intel(R) Core(TM) i7-7800X CPU @ 3.50GHz on Linux-5.3.0-19-generic-x86\_64-with-debian-buster-sid}\label{fig:scaling}}
 \end{figure*}

Results are summarized in Figure~\ref{fig:scaling}, where we plot the average time as a function of $N\cdot \log(N)$. 
It can be seen that the clustering time for 250000 points is about 15, 5, and 20 min respectively for $d$ equals to 2, 4, and 8, so at relatively high dimensions the run time is of the order of a fraction of an hour. 
The aforementioned times are obtained on an Intel(R) Core(TM) i7-7800X CPU @3.50GHz using the Python implementation
of DPA (see Code availability Section). This implementation heavily relies on the Scikit-learn~\cite{scikit-learn} Nearest Neighbors search, currently implemented with  library that uses a heuristic selecting between brute-force, ball-tree~\cite{balltree} and kd-tree~\cite{kd-tree} algorithms. The scaling of the algorithm for large values of $N$ is $N\cdot \log(N)$, as shown by the trends seen in Figure~\ref{fig:scaling}.
The scaling with the dimensionality $d$ is more complex, because an increased value for $d$ usually corresponds to a reduction in the optimal neighborhood size $\hat{k}$ around each point. Therefore, while the computational cost of the Nearest Neighbor search increases with $d$, the steps in the algorithm depending on the value of $\hat{k}$ show a better performance.  This explains why the run times estimated at $d=4$ are lower than those at $d=2$. However, this behavior depends on the specific characteristics of the data set, and it is hard to generalize.

\section{Conclusion}

In this work we introduce a tool to analyze large and multidimensional data sets{, which} is specifically designed to
treat cases in which the standard projection technique (e.g. PCA) gives poor results.

Data sets can often be described as realizations of an underlying probability distribution, whose density has support 
in the space of the features (coordinates) of the data. Therefore, the idea is to reconstruct the \emph{topography} of  the probability distribution
from which the data are generated. Our clusters topography is a list of probability peaks, each characterized by its properties: the height of the probability maximum, the population, the list of neighboring peaks, etc. {Our tool employs only the distance between points, avoiding the use of any coordinate system or projection, and its derivation is based on three assumptions:} {(i) the data points are generated independently from a probability distribution function; (ii) this probability distribution is continuous and (iii) the distance between data points is a metric, and in particular it should  satisfy the triangle inequality.}

We {illustrated} two ways to visualize this topography: a hierarchical representation equivalent to the one used in refs.~\cite{hartigan1981consistency,HDBSCAN,RobustSL,RobustDB} and a graph representation tantamount to the one of Markov State Models~\cite{MSM2010}.  Of course, one can imagine other graphical representations.

The topography  is reconstructed by  a modified version of the {unsupervised} Density Peaks clustering algorithm~\cite{AlexAle} {whose key ingredient is} the PA$k$ density estimator. {The PA$k$ estimator is parameter free and, with respect to other non-parametric estimators,} provides an accurate estimation of the probability density at the data points in the manifold {in which they lay, and not in the space of coordinates.}  {It also {provides} the error associated with the density estimate: this is crucial to assess} the statistical significance of the peaks found by the clustering procedure and, as a consequence, to discriminate between real features of the underlying probability distribution function and artifacts due to finite sampling.
The use of PA$k$ coupled with the Density Peaks clustering algorithm thus leads to an accurate detection of the main features of the underlying probability distribution, where the statistical reliability of the probability peaks is quantified by the $Z$ score measuring the negative of the probability that the peak is generated by a statistical fluctuation. By choosing a threshold on $Z$, one can filter out the peaks that are less reliable and obtain a more compact representation of the data. 
{A drawback of the method is that, due to the curse of dimensionality, the PA$k$ algorithm does not work well when the intrinsic dimension of the data set is higher than 10-20~\cite{FEwithoutCV}, and therefore neither will DPA. However, many important data sets lay in a manifold that can be twisted and topologically complex but whose intrinsic dimension is typically much smaller than the number of coordinates of the system~\cite{Bickel2005}, which is instead typically very large.} 
The method performs well in the two dimensional toy examples in Figure~\ref{fig:Clusterexpl} and~\ref{fig:spirali}. {The artificial low-dimensional data sets AGG, SPIR3 and HORSE are obtained respectively from the Aggregation~\cite{Aggregation}, Spirals~\cite{Spiral} and Jain~\cite{Jain} data set, well-known as test sets for investigating density-based clustering algorithms.
In ref.~\cite{JIANG2019702} for example Aggregation, Spirals and Jain are used to test the performance of the algorithms on irregular-shaped clusters or clusters of varying sizes. However, real-world data sets are usually noisy, and background noise is not present in those data sets. Moreover, if optimized on data sets with well-separated clusters, algorithms may be less performant on real-world data sets where clusters are typically connected by regions with a lower density of data points, as in the case of ref.~\cite{wang2018density}. Therefore, for our experiments we generated AGG, SPIR3, and HORSE using a probability distribution function built as a sum of Gaussian functions centered in the original data points. The resulting data sets maintain the same number of clusters and dimensionality of the original ones, with the addition of noise points to better resemble the continuous distribution of data points typical of real-world data sets.}
Furthermore, we investigated the method on handwritten numbers (NMIST) and the PUA clan {with intrinsic dimension of 8 and 9 respectively, as estimated by the TWO-NN method~\cite{faccoID}. The MNIST hand-written digits data set is largely
used in Machine Learning to benchmark density-based clustering methods, although the presence of different handwritings style and the high dimensionality of the data set can make the correct identification of ciphers challenging, as shown in ref.~\cite{hess2019spectacl}.
The PUA data set has been recently analyzed also in ref.~\cite{russo2020dpcfam} for automatic protein domain classification.}

The method shows its real power when the number of features is huge, as in the cases of MNIST and the PUA clan where computing the density in the manifold in which the data lay, instead of computing it in the coordinate  space, is the key for a successful reconstruction of the topography. Moreover, the knowledge of the density at the borders permits the visualization of the relationship between these modes --the topography-- in several ways providing a visual grasp of the structure of the data set with an unprecedented level of detail.

\section*{Code availability}
{
A Python implementation of the algorithm is available  on github, at https://github.com/mariaderrico/DPA \label{codice_python}. It is also available a Fortran implementation at: https://github.com/alexdepremia/Advanced-Density-Peaks.\label{codice_fortran} 
}

\section*{Acknowledgment}
The authors warmly acknowledge Michele Allegra, Marco Borelli, Giovanni Pinamonti, Daniele Amati and Marco Punta for several useful discussions.

\section*{Author contributions statement}

\textbf{Maria d'Errico and Alex Rodriguez:} Conceptualization, Methodology, Investigation, Software, Validation, Formal analysis, Visualization, Writing- Original draft preparation, Writing- Reviewing and Editing. \textbf{Elena Facco:} Data curation, Formal analysis, Visualization, Writing- Original draft preparation, Writing- Reviewing and Editing. \textbf{Alessandro Laio:} Conceptualization, Methodology, Investigation, Formal analysis, Visualization, Writing- Original draft preparation, Writing- Reviewing and Editing.

\section*{Additional information}

A document with the Supplementary Information mentioned in the manuscript (SI.pdf) is provided along with the coordinates and ground-truth classifications employed for computing the Normalized Mutual Information values provided in Figure~\ref{fig:performance} ($X$.txt \& $X$\_gt.txt, with $X$= CLUS8, SPIR2, AGG, SPIR3, HORSE).

\section*{Competing interest statement}

The authors declare that they have no competing interests.

\bibliographystyle{elsarticle-harv}
\bibliography{no_doi_no_url.bib}

\begin{thebibliography}{50}
\expandafter\ifx\csname natexlab\endcsname\relax\def\natexlab#1{#1}\fi
\providecommand{\url}[1]{\texttt{#1}}
\providecommand{\href}[2]{#2}
\providecommand{\path}[1]{#1}
\providecommand{\DOIprefix}{doi:}
\providecommand{\ArXivprefix}{arXiv:}
\providecommand{\URLprefix}{URL: }
\providecommand{\Pubmedprefix}{pmid:}
\providecommand{\doi}[1]{\href{http://dx.doi.org/#1}{\path{#1}}}
\providecommand{\Pubmed}[1]{\href{pmid:#1}{\path{#1}}}
\providecommand{\bibinfo}[2]{#2}
\ifx\xfnm\relax \def\xfnm[#1]{\unskip,\space#1}\fi
\bibitem[{Bentley(1975)}]{kd-tree}
\bibinfo{author}{Bentley, J.L.}, \bibinfo{year}{1975}.
\newblock \bibinfo{title}{Multidimensional binary search trees used for
  associative searching}.
\newblock \bibinfo{journal}{Commun. ACM} \bibinfo{volume}{18},
  \bibinfo{pages}{509–517}.
\bibitem[{Blei et~al.(2006)Blei, Jordan et~al.}]{DPGMM}
\bibinfo{author}{Blei, D.M.}, \bibinfo{author}{Jordan, M.I.}, et~al.,
  \bibinfo{year}{2006}.
\newblock \bibinfo{title}{Variational inference for dirichlet process
  mixtures}.
\newblock \bibinfo{journal}{Bayesian analysis} \bibinfo{volume}{1},
  \bibinfo{pages}{121--143}.
\bibitem[{Bunte et~al.(2012)Bunte, Biehl and Hammer}]{bunte2012general}
\bibinfo{author}{Bunte, K.}, \bibinfo{author}{Biehl, M.},
  \bibinfo{author}{Hammer, B.}, \bibinfo{year}{2012}.
\newblock \bibinfo{title}{A general framework for dimensionality-reducing data
  visualization mapping}.
\newblock \bibinfo{journal}{Neural Computation} \bibinfo{volume}{24},
  \bibinfo{pages}{771--804}.
\bibitem[{Camastra and Staiano(2016)}]{CAMASTRA201626}
\bibinfo{author}{Camastra, F.}, \bibinfo{author}{Staiano, A.},
  \bibinfo{year}{2016}.
\newblock \bibinfo{title}{Intrinsic dimension estimation: Advances and open
  problems}.
\newblock \bibinfo{journal}{Information Sciences} \bibinfo{volume}{328},
  \bibinfo{pages}{26 -- 41}.
\bibitem[{Campello et~al.(2013)Campello, Moulavi and Sander}]{HDBSCAN}
\bibinfo{author}{Campello, R.J.}, \bibinfo{author}{Moulavi, D.},
  \bibinfo{author}{Sander, J.}, \bibinfo{year}{2013}.
\newblock \bibinfo{title}{Density-based clustering based on hierarchical
  density estimates}, in: \bibinfo{booktitle}{Pacific-Asia Conference on
  Knowledge Discovery and Data Mining}, \bibinfo{organization}{Springer}. pp.
  \bibinfo{pages}{160--172}.
\bibitem[{Ceriotti et~al.(2011)Ceriotti, Tribello and Parrinello}]{SKETCHMAP}
\bibinfo{author}{Ceriotti, M.}, \bibinfo{author}{Tribello, G.A.},
  \bibinfo{author}{Parrinello, M.}, \bibinfo{year}{2011}.
\newblock \bibinfo{title}{Simplifying the representation of complex free-energy
  landscapes using sketch-map}.
\newblock \bibinfo{journal}{Proceedings of the National Academy of Sciences}
  \bibinfo{volume}{108}, \bibinfo{pages}{13023--13028}.
\bibitem[{Chang and Yeung(2008)}]{Spiral}
\bibinfo{author}{Chang, H.}, \bibinfo{author}{Yeung, D.Y.},
  \bibinfo{year}{2008}.
\newblock \bibinfo{title}{Robust path-based spectral clustering}.
\newblock \bibinfo{journal}{Pattern Recognition} \bibinfo{volume}{41},
  \bibinfo{pages}{191--203}.
\bibitem[{Chaudhuri et~al.(2014)Chaudhuri, Dasgupta, Kpotufe and von
  Luxburg}]{RobustSL}
\bibinfo{author}{Chaudhuri, K.}, \bibinfo{author}{Dasgupta, S.},
  \bibinfo{author}{Kpotufe, S.}, \bibinfo{author}{von Luxburg, U.},
  \bibinfo{year}{2014}.
\newblock \bibinfo{title}{Consistent procedures for cluster tree estimation and
  pruning}.
\newblock \bibinfo{journal}{IEEE Transactions on Information Theory}
  \bibinfo{volume}{60}, \bibinfo{pages}{7900--7912}.
\bibitem[{Coifman et~al.(2005)Coifman, Lafon, Lee, Maggioni, Nadler, Warner and
  Zucker}]{DM}
\bibinfo{author}{Coifman, R.R.}, \bibinfo{author}{Lafon, S.},
  \bibinfo{author}{Lee, A.B.}, \bibinfo{author}{Maggioni, M.},
  \bibinfo{author}{Nadler, B.}, \bibinfo{author}{Warner, F.},
  \bibinfo{author}{Zucker, S.W.}, \bibinfo{year}{2005}.
\newblock \bibinfo{title}{Geometric diffusions as a tool for harmonic analysis
  and structure definition of data: Diffusion maps}.
\newblock \bibinfo{journal}{Proceedings of the National Academy of Sciences of
  the United States of America} \bibinfo{volume}{102},
  \bibinfo{pages}{7426--7431}.
\bibitem[{Comaniciu and Meer(2002)}]{MeanShift2004}
\bibinfo{author}{Comaniciu, D.}, \bibinfo{author}{Meer, P.},
  \bibinfo{year}{2002}.
\newblock \bibinfo{title}{Mean shift: A robust approach toward feature space
  analysis}.
\newblock \bibinfo{journal}{IEEE Transactions on Pattern Analysis and Machine
  Intelligence} \bibinfo{volume}{24}, \bibinfo{pages}{603--619}.
\bibitem[{Ester et~al.(1996)Ester, Kriegel, Sander and Xu}]{corrected_DBSCAN}
\bibinfo{author}{Ester, M.}, \bibinfo{author}{Kriegel, H.P.},
  \bibinfo{author}{Sander, J.}, \bibinfo{author}{Xu, X.}, \bibinfo{year}{1996}.
\newblock \bibinfo{title}{A density-based algorithm for discovering clusters in
  large spatial databases with noise}, in: \bibinfo{booktitle}{Proceedings of
  the Second International Conference on Knowledge Discovery and Data Mining},
  \bibinfo{publisher}{AAAI Press}. p. \bibinfo{pages}{226–231}.
\bibitem[{Facco et~al.(2017)Facco, d'Errico, Rodriguez and Laio}]{faccoID}
\bibinfo{author}{Facco, E.}, \bibinfo{author}{d'Errico, M.},
  \bibinfo{author}{Rodriguez, A.}, \bibinfo{author}{Laio, A.},
  \bibinfo{year}{2017}.
\newblock \bibinfo{title}{Estimating the intrinsic dimension of datasets by a
  minimal neighborhood information.}
\newblock \bibinfo{journal}{Scientific reports} \bibinfo{volume}{7},
  \bibinfo{pages}{12140}.
\bibitem[{Facco et~al.(2019)Facco, Pagnani, Russo and Laio}]{IDprot}
\bibinfo{author}{Facco, E.}, \bibinfo{author}{Pagnani, A.},
  \bibinfo{author}{Russo, E.T.}, \bibinfo{author}{Laio, A.},
  \bibinfo{year}{2019}.
\newblock \bibinfo{title}{The intrinsic dimension of protein sequence
  evolution}.
\newblock \bibinfo{journal}{PLOS Computational Biology} \bibinfo{volume}{15},
  \bibinfo{pages}{1--16}.
\bibitem[{Finn et~al.(2014)Finn, Bateman, Clements, Coggill, Eberhardt, Eddy,
  Heger, Hetherington, Holm, Mistry, Sonnhammer, Tate and Punta}]{PFAM}
\bibinfo{author}{Finn, R.D.}, \bibinfo{author}{Bateman, A.},
  \bibinfo{author}{Clements, J.}, \bibinfo{author}{Coggill, P.},
  \bibinfo{author}{Eberhardt, R.Y.}, \bibinfo{author}{Eddy, S.R.},
  \bibinfo{author}{Heger, A.}, \bibinfo{author}{Hetherington, K.},
  \bibinfo{author}{Holm, L.}, \bibinfo{author}{Mistry, J.},
  \bibinfo{author}{Sonnhammer, E.L.L.}, \bibinfo{author}{Tate, J.},
  \bibinfo{author}{Punta, M.}, \bibinfo{year}{2014}.
\newblock \bibinfo{title}{{Pfam: the protein families database}}.
\newblock \bibinfo{journal}{Nucleic Acids Research} \bibinfo{volume}{42},
  \bibinfo{pages}{D222--D230}.
\bibitem[{Finn et~al.(2016)Finn, Coggill, Eberhardt, Eddy, Mistry, Mitchell,
  Potter, Punta, Qureshi, Sangrador-Vegas, Salazar, Tate and
  Bateman}]{PFAMlatest}
\bibinfo{author}{Finn, R.D.}, \bibinfo{author}{Coggill, P.},
  \bibinfo{author}{Eberhardt, R.Y.}, \bibinfo{author}{Eddy, S.R.},
  \bibinfo{author}{Mistry, J.}, \bibinfo{author}{Mitchell, A.L.},
  \bibinfo{author}{Potter, S.C.}, \bibinfo{author}{Punta, M.},
  \bibinfo{author}{Qureshi, M.}, \bibinfo{author}{Sangrador-Vegas, A.},
  \bibinfo{author}{Salazar, G.A.}, \bibinfo{author}{Tate, J.},
  \bibinfo{author}{Bateman, A.}, \bibinfo{year}{2016}.
\newblock \bibinfo{title}{{The Pfam protein families database: towards a more
  sustainable future}}.
\newblock \bibinfo{journal}{Nucleic Acids Research} \bibinfo{volume}{44},
  \bibinfo{pages}{D279--D285}.
\bibitem[{Gionis et~al.(2007)Gionis, Mannila and Tsaparas}]{Aggregation}
\bibinfo{author}{Gionis, A.}, \bibinfo{author}{Mannila, H.},
  \bibinfo{author}{Tsaparas, P.}, \bibinfo{year}{2007}.
\newblock \bibinfo{title}{Clustering aggregation}.
\newblock \bibinfo{journal}{ACM Transactions on Knowledge Discovery from Data
  (TKDD)} \bibinfo{volume}{1}, \bibinfo{pages}{4}.
\bibitem[{Gisbrecht and Hammer(2015)}]{gisbrecht2015data}
\bibinfo{author}{Gisbrecht, A.}, \bibinfo{author}{Hammer, B.},
  \bibinfo{year}{2015}.
\newblock \bibinfo{title}{Data visualization by nonlinear dimensionality
  reduction}.
\newblock \bibinfo{journal}{Wiley Interdisciplinary Reviews: Data Mining and
  Knowledge Discovery} \bibinfo{volume}{5}, \bibinfo{pages}{51--73}.
\bibitem[{Granata and Carnevale(2016)}]{granata2016accurate}
\bibinfo{author}{Granata, D.}, \bibinfo{author}{Carnevale, V.},
  \bibinfo{year}{2016}.
\newblock \bibinfo{title}{Accurate estimation of the intrinsic dimension using
  graph distances: Unraveling the geometric complexity of datasets}.
\newblock \bibinfo{journal}{Scientific reports} \bibinfo{volume}{6},
  \bibinfo{pages}{31377}.
\bibitem[{Hartigan(1981a)}]{hartigan1981consistency}
\bibinfo{author}{Hartigan, J.A.}, \bibinfo{year}{1981}a.
\newblock \bibinfo{title}{Consistency of single linkage for high-density
  clusters}.
\newblock \bibinfo{journal}{Journal of the American Statistical Association}
  \bibinfo{volume}{76}, \bibinfo{pages}{388--394}.
\bibitem[{Hartigan(1981b)}]{SingleLinkage}
\bibinfo{author}{Hartigan, J.A.}, \bibinfo{year}{1981}b.
\newblock \bibinfo{title}{Consistency of single linkage for high-density
  clusters}.
\newblock \bibinfo{journal}{Journal of the American Statistical Association}
  \bibinfo{volume}{76}, \bibinfo{pages}{388--394}.
\bibitem[{Hess et~al.(2019)Hess, Duivesteijn, Honysz and
  Morik}]{hess2019spectacl}
\bibinfo{author}{Hess, S.}, \bibinfo{author}{Duivesteijn, W.},
  \bibinfo{author}{Honysz, P.}, \bibinfo{author}{Morik, K.},
  \bibinfo{year}{2019}.
\newblock \bibinfo{title}{The spectacl of nonconvex clustering: a spectral
  approach to density-based clustering}, in: \bibinfo{booktitle}{Proceedings of
  the AAAI Conference on Artificial Intelligence}, pp.
  \bibinfo{pages}{3788--3795}.
\bibitem[{Jain and Law(2005)}]{Jain}
\bibinfo{author}{Jain, A.K.}, \bibinfo{author}{Law, M.H.},
  \bibinfo{year}{2005}.
\newblock \bibinfo{title}{Data clustering: A user’s dilemma}, in:
  \bibinfo{booktitle}{International conference on pattern recognition and
  machine intelligence}, \bibinfo{organization}{Springer}. pp.
  \bibinfo{pages}{1--10}.
\bibitem[{Jiang et~al.(2019)Jiang, Chen, Meng, Wang and Li}]{JIANG2019702}
\bibinfo{author}{Jiang, J.}, \bibinfo{author}{Chen, Y.}, \bibinfo{author}{Meng,
  X.}, \bibinfo{author}{Wang, L.}, \bibinfo{author}{Li, K.},
  \bibinfo{year}{2019}.
\newblock \bibinfo{title}{A novel density peaks clustering algorithm based on k
  nearest neighbors for improving assignment process}.
\newblock \bibinfo{journal}{Physica A: Statistical Mechanics and its
  Applications} \bibinfo{volume}{523}, \bibinfo{pages}{702 -- 713}.
\bibitem[{LeCun et~al.(1998)LeCun, Bottou, Bengio and Haffner}]{MNIST}
\bibinfo{author}{LeCun, Y.}, \bibinfo{author}{Bottou, L.},
  \bibinfo{author}{Bengio, Y.}, \bibinfo{author}{Haffner, P.},
  \bibinfo{year}{1998}.
\newblock \bibinfo{title}{Gradient-based learning applied to document
  recognition}.
\newblock \bibinfo{journal}{Proceedings of the IEEE} \bibinfo{volume}{86},
  \bibinfo{pages}{2278--2324}.
\bibitem[{Levina and Bickel(2005)}]{Bickel2005}
\bibinfo{author}{Levina, E.}, \bibinfo{author}{Bickel, P.J.},
  \bibinfo{year}{2005}.
\newblock \bibinfo{title}{Maximum likelihood estimation of intrinsic
  dimension}, in: \bibinfo{editor}{Saul, L.K.}, \bibinfo{editor}{Weiss, Y.},
  \bibinfo{editor}{Bottou, L.} (Eds.), \bibinfo{booktitle}{Advances in Neural
  Information Processing Systems 17}. \bibinfo{publisher}{MIT Press}, pp.
  \bibinfo{pages}{777--784}.
\bibitem[{Liang and Chen(2016)}]{LIANG201652}
\bibinfo{author}{Liang, Z.}, \bibinfo{author}{Chen, P.}, \bibinfo{year}{2016}.
\newblock \bibinfo{title}{Delta-density based clustering with a
  divide-and-conquer strategy: 3dc clustering}.
\newblock \bibinfo{journal}{Pattern Recognition Letters} \bibinfo{volume}{73},
  \bibinfo{pages}{52 -- 59}.
\bibitem[{Maaten and Hinton(2008)}]{tsne}
\bibinfo{author}{Maaten, L.v.d.}, \bibinfo{author}{Hinton, G.},
  \bibinfo{year}{2008}.
\newblock \bibinfo{title}{Visualizing data using t-sne}.
\newblock \bibinfo{journal}{Journal of machine learning research}
  \bibinfo{volume}{9}, \bibinfo{pages}{2579--2605}.
\bibitem[{McInnes et~al.(2017)McInnes, Healy and Astels}]{hdbscan-python}
\bibinfo{author}{McInnes, L.}, \bibinfo{author}{Healy, J.},
  \bibinfo{author}{Astels, S.}, \bibinfo{year}{2017}.
\newblock \bibinfo{title}{hdbscan: Hierarchical density based clustering}.
\newblock \bibinfo{journal}{The Journal of Open Source Software}
  \bibinfo{volume}{2}.
\bibitem[{Mehmood et~al.(2016)Mehmood, Zhang, Bie, Dawood and Ahmad}]{MEHMOOD}
\bibinfo{author}{Mehmood, R.}, \bibinfo{author}{Zhang, G.},
  \bibinfo{author}{Bie, R.}, \bibinfo{author}{Dawood, H.},
  \bibinfo{author}{Ahmad, H.}, \bibinfo{year}{2016}.
\newblock \bibinfo{title}{Clustering by fast search and find of density peaks
  via heat diffusion}.
\newblock \bibinfo{journal}{Neurocomputing} \bibinfo{volume}{208},
  \bibinfo{pages}{210 -- 217}.
\newblock \bibinfo{note}{SI: BridgingSemantic}.
\bibitem[{Minnotte(1997)}]{minnotte1997nonparametric}
\bibinfo{author}{Minnotte, M.C.}, \bibinfo{year}{1997}.
\newblock \bibinfo{title}{Nonparametric testing of the existence of modes}.
\newblock \bibinfo{journal}{The Annals of Statistics} ,
  \bibinfo{pages}{1646--1660}.
\bibitem[{Neyman and Pearson(1933)}]{LikeRat}
\bibinfo{author}{Neyman, J.}, \bibinfo{author}{Pearson, E.S.},
  \bibinfo{year}{1933}.
\newblock \bibinfo{title}{On the problem of the most efficient tests of
  statistical hypotheses}.
\newblock \bibinfo{journal}{Philosophical Transactions of the Royal Society of
  London. Series A, Containing Papers of a Mathematical or Physical Character}
  \bibinfo{volume}{231}, \bibinfo{pages}{289--337}.
\bibitem[{Ng et~al.(2002)Ng, Jordan and Weiss}]{ng2002spectral}
\bibinfo{author}{Ng, A.Y.}, \bibinfo{author}{Jordan, M.I.},
  \bibinfo{author}{Weiss, Y.}, \bibinfo{year}{2002}.
\newblock \bibinfo{title}{On spectral clustering: Analysis and an algorithm},
  in: \bibinfo{booktitle}{Advances in neural information processing systems},
  pp. \bibinfo{pages}{849--856}.
\bibitem[{Omohundro(1989)}]{balltree}
\bibinfo{author}{Omohundro, S.M.}, \bibinfo{year}{1989}.
\newblock \bibinfo{title}{Five balltree construction algorithms}.
\newblock \bibinfo{publisher}{International Computer Science Institute
  Berkeley}.
\bibitem[{Pande et~al.(2010)Pande, Beauchamp and Bowman}]{MSM2010}
\bibinfo{author}{Pande, V.S.}, \bibinfo{author}{Beauchamp, K.},
  \bibinfo{author}{Bowman, G.R.}, \bibinfo{year}{2010}.
\newblock \bibinfo{title}{Everything you wanted to know about markov state
  models but were afraid to ask}.
\newblock \bibinfo{journal}{Methods} \bibinfo{volume}{52}, \bibinfo{pages}{99
  -- 105}.
\newblock \bibinfo{note}{Protein Folding}.
\bibitem[{Pedregosa et~al.(2011)Pedregosa, Varoquaux, Gramfort, Michel,
  Thirion, Grisel, Blondel, Prettenhofer, Weiss, Dubourg, Vanderplas, Passos,
  Cournapeau, Brucher, Perrot and Duchesnay}]{scikit-learn}
\bibinfo{author}{Pedregosa, F.}, \bibinfo{author}{Varoquaux, G.},
  \bibinfo{author}{Gramfort, A.}, \bibinfo{author}{Michel, V.},
  \bibinfo{author}{Thirion, B.}, \bibinfo{author}{Grisel, O.},
  \bibinfo{author}{Blondel, M.}, \bibinfo{author}{Prettenhofer, P.},
  \bibinfo{author}{Weiss, R.}, \bibinfo{author}{Dubourg, V.},
  \bibinfo{author}{Vanderplas, J.}, \bibinfo{author}{Passos, A.},
  \bibinfo{author}{Cournapeau, D.}, \bibinfo{author}{Brucher, M.},
  \bibinfo{author}{Perrot, M.}, \bibinfo{author}{Duchesnay, E.},
  \bibinfo{year}{2011}.
\newblock \bibinfo{title}{Scikit-learn: Machine learning in {P}ython}.
\newblock \bibinfo{journal}{Journal of Machine Learning Research}
  \bibinfo{volume}{12}, \bibinfo{pages}{2825--2830}.
\bibitem[{Ringn{\'e}r(2008)}]{PCA}
\bibinfo{author}{Ringn{\'e}r, M.}, \bibinfo{year}{2008}.
\newblock \bibinfo{title}{What is principal component analysis?}
\newblock \bibinfo{journal}{Nature biotechnology} \bibinfo{volume}{26},
  \bibinfo{pages}{303--304}.
\bibitem[{Rodriguez et~al.(2018)Rodriguez, d’Errico, Facco and
  Laio}]{FEwithoutCV}
\bibinfo{author}{Rodriguez, A.}, \bibinfo{author}{d’Errico, M.},
  \bibinfo{author}{Facco, E.}, \bibinfo{author}{Laio, A.},
  \bibinfo{year}{2018}.
\newblock \bibinfo{title}{Computing the free energy without collective
  variables}.
\newblock \bibinfo{journal}{Journal of chemical theory and computation}
  \bibinfo{volume}{14}, \bibinfo{pages}{1206--1215}.
\bibitem[{Rodriguez and Laio(2014)}]{AlexAle}
\bibinfo{author}{Rodriguez, A.}, \bibinfo{author}{Laio, A.},
  \bibinfo{year}{2014}.
\newblock \bibinfo{title}{Clustering by fast search and find of density peaks}.
\newblock \bibinfo{journal}{Science} \bibinfo{volume}{344},
  \bibinfo{pages}{1492--1496}.
\bibitem[{Roweis and Saul(2000)}]{LLE}
\bibinfo{author}{Roweis, S.T.}, \bibinfo{author}{Saul, L.K.},
  \bibinfo{year}{2000}.
\newblock \bibinfo{title}{Nonlinear dimensionality reduction by locally linear
  embedding}.
\newblock \bibinfo{journal}{Science} \bibinfo{volume}{290},
  \bibinfo{pages}{2323--2326}.
\bibitem[{Russo et~al.(2020)Russo, Laio and Punta}]{russo2020dpcfam}
\bibinfo{author}{Russo, E.T.}, \bibinfo{author}{Laio, A.},
  \bibinfo{author}{Punta, M.}, \bibinfo{year}{2020}.
\newblock \bibinfo{title}{Dpcfam: a new method for unsupervised protein family
  classification}.
\newblock \bibinfo{journal}{bioRxiv} .
\bibitem[{Shieh et~al.(2011)Shieh, Hashimoto and Airoldi}]{TPE}
\bibinfo{author}{Shieh, A.D.}, \bibinfo{author}{Hashimoto, T.B.},
  \bibinfo{author}{Airoldi, E.M.}, \bibinfo{year}{2011}.
\newblock \bibinfo{title}{Tree preserving embedding}.
\newblock \bibinfo{journal}{Proceedings of the National Academy of Sciences}
  \bibinfo{volume}{108}, \bibinfo{pages}{16916--16921}.
\bibitem[{Silverman(1981)}]{silverman1981using}
\bibinfo{author}{Silverman, B.W.}, \bibinfo{year}{1981}.
\newblock \bibinfo{title}{Using kernel density estimates to investigate
  multimodality}.
\newblock \bibinfo{journal}{Journal of the Royal Statistical Society. Series B
  (Methodological)} , \bibinfo{pages}{97--99}.
\bibitem[{Simard et~al.(1993)Simard, LeCun and Denker}]{TanDist}
\bibinfo{author}{Simard, P.}, \bibinfo{author}{LeCun, Y.},
  \bibinfo{author}{Denker, J.S.}, \bibinfo{year}{1993}.
\newblock \bibinfo{title}{Efficient pattern recognition using a new
  transformation distance}, in: \bibinfo{booktitle}{Advances in neural
  information processing systems}, pp. \bibinfo{pages}{50--58}.
\bibitem[{Sittel and Stock(2016)}]{RobustDB}
\bibinfo{author}{Sittel, F.}, \bibinfo{author}{Stock, G.},
  \bibinfo{year}{2016}.
\newblock \bibinfo{title}{Robust density-based clustering to identify
  metastable conformational states of proteins}.
\newblock \bibinfo{journal}{Journal of chemical theory and computation}
  \bibinfo{volume}{12}, \bibinfo{pages}{2426--2435}.
\bibitem[{Sormani et~al.(2019)Sormani, Rodriguez and
  Laio}]{sormani2019explicit}
\bibinfo{author}{Sormani, G.}, \bibinfo{author}{Rodriguez, A.},
  \bibinfo{author}{Laio, A.}, \bibinfo{year}{2019}.
\newblock \bibinfo{title}{Explicit characterization of the free-energy
  landscape of a protein in the space of all its c$\alpha$ carbons}.
\newblock \bibinfo{journal}{Journal of Chemical Theory and Computation}
  \bibinfo{volume}{16}, \bibinfo{pages}{80--87}.
\bibitem[{Tenenbaum et~al.(2000)Tenenbaum, De~Silva and
  Langford}]{tenenbaum2000global}
\bibinfo{author}{Tenenbaum, J.B.}, \bibinfo{author}{De~Silva, V.},
  \bibinfo{author}{Langford, J.C.}, \bibinfo{year}{2000}.
\newblock \bibinfo{title}{A global geometric framework for nonlinear
  dimensionality reduction}.
\newblock \bibinfo{journal}{science} \bibinfo{volume}{290},
  \bibinfo{pages}{2319--2323}.
\bibitem[{Torgerson(1952)}]{torgerson1952multidimensional}
\bibinfo{author}{Torgerson, W.S.}, \bibinfo{year}{1952}.
\newblock \bibinfo{title}{Multidimensional scaling: I. theory and method}.
\newblock \bibinfo{journal}{Psychometrika} \bibinfo{volume}{17},
  \bibinfo{pages}{401--419}.
\bibitem[{Vinh et~al.(2010)Vinh, Epps and Bailey}]{AMI}
\bibinfo{author}{Vinh, N.X.}, \bibinfo{author}{Epps, J.},
  \bibinfo{author}{Bailey, J.}, \bibinfo{year}{2010}.
\newblock \bibinfo{title}{Information theoretic measures for clusterings
  comparison: Variants, properties, normalization and correction for chance}.
\newblock \bibinfo{journal}{J. Mach. Learn. Res.} \bibinfo{volume}{11},
  \bibinfo{pages}{2837--2854}.
\bibitem[{Wang et~al.(2018)Wang, Pang and Zhou}]{wang2018density}
\bibinfo{author}{Wang, Y.}, \bibinfo{author}{Pang, W.}, \bibinfo{author}{Zhou,
  Y.}, \bibinfo{year}{2018}.
\newblock \bibinfo{title}{Density propagation based adaptive multi-density
  clustering algorithm}.
\newblock \bibinfo{journal}{Plos one} \bibinfo{volume}{13},
  \bibinfo{pages}{e0198948}.
\bibitem[{Xu and Tian(2015)}]{Xu2015}
\bibinfo{author}{Xu, D.}, \bibinfo{author}{Tian, Y.}, \bibinfo{year}{2015}.
\newblock \bibinfo{title}{A comprehensive survey of clustering algorithms}.
\newblock \bibinfo{journal}{Annals of Data Science} \bibinfo{volume}{2},
  \bibinfo{pages}{165--193}.

\end{thebibliography}







\end{document}